\documentclass[nohyperref]{article}

\usepackage{microtype}
\usepackage{graphicx}
\usepackage{subfigure}
\usepackage{booktabs} % for professional tables

\usepackage{hyperref}

\usepackage[accepted]{icml2023}

\usepackage{amsmath}
\usepackage{amssymb}
\usepackage{mathtools}
\usepackage{amsthm}

\usepackage[capitalize,noabbrev]{cleveref}

\theoremstyle{plain}

\theoremstyle{definition}

\theoremstyle{remark}

\usepackage[textsize=tiny]{todonotes}

\usepackage{wrapfig}				%
\usepackage{multicol}				%
\usepackage{multirow} 				%
\usepackage{makecell}

\newcommand{\bs}{\mathbf{s}}		%
\newcommand{\ba}{\mathbf{a}}		%
\newcommand{\bz}{\mathbf{z}}		%
\newcommand{\loss}{\ell}			%
\newcommand{\eg}{\textit{e.g.},\ }	%
\newcommand{\ie}{\textit{i.e.},\ }	%

\begin{document}

\twocolumn[
\icmltitle{Beyond Reward: Offline Preference-guided Policy Optimization}



\icmlsetsymbol{equal}{*}

\begin{icmlauthorlist}
\icmlauthor{Yachen Kang}{ZJU,WU}
\icmlauthor{Diyuan Shi}{WU}
\icmlauthor{Jinxin Liu}{WU}
\icmlauthor{Li He}{WU}
\icmlauthor{Donglin Wang}{WU}
\end{icmlauthorlist}

\icmlaffiliation{ZJU}{College of Computer Science and Technology, Zhejiang University, Hangzhou, Zhejiang, China}
\icmlaffiliation{WU}{Machine Intelligence Lab (MiLAB) of the School of Engineering, Westlake University, Hangzhou, Zhejiang, China}

\icmlcorrespondingauthor{Donglin Wang}{wangdonglin@westlake.edu.cn}
\icmlcorrespondingauthor{Yachen Kang}{kangyachen@westlake.edu.cn}

\icmlkeywords{Machine Learning, ICML}

\vskip 0.3in
]



\printAffiliationsAndNotice{}  % leave blank if no need to mention equal contribution

\begin{abstract}
This study focuses on the topic of offline preference-based reinforcement learning (PbRL), a variant of conventional reinforcement learning that dispenses with the need for online interaction or specification of reward functions.
Instead, the agent is provided with fixed offline trajectories and human preferences between pairs of trajectories to extract the dynamics and task information, respectively.
Since the dynamics and task information are orthogonal, a naive approach would involve using preference-based reward learning followed by an off-the-shelf offline RL algorithm.
However, this requires the separate learning of a scalar reward function, which is assumed to be an information bottleneck of the learning process.
To address this issue, we propose the offline preference-guided policy optimization (OPPO) paradigm, which models offline trajectories and preferences in a one-step process, eliminating the need for separately learning a reward function.
OPPO achieves this by introducing an offline hindsight information matching objective for optimizing a contextual policy and a preference modeling objective for finding the optimal context.
OPPO further integrates a well-performing decision policy by optimizing the two objectives iteratively.
Our empirical results demonstrate that OPPO effectively models offline preferences and outperforms prior competing baselines, including offline RL algorithms performed over either true or pseudo reward function specifications.
Our code is available on the project website: \textcolor{magenta}{\url{https://sites.google.com/view/oppo-icml-2023}}.
\end{abstract}

\section{Introduction}
Deep reinforcement learning (RL) offers a versatile framework for acquiring task-oriented behaviors, as evidenced by a growing body of literature~\citep{NateKohl2004PolicyGR,JensKober2008PolicySF,JensKober2013ReinforcementLI,DavidSilver2017MasteringTG,DmitryKalashnikov2018QTOptSD,OriolVinyals2019GrandmasterLI}.
In this framework, the "task" is frequently expressed as maximizing the cumulative reward of trajectories produced by deploying the learning policy in the corresponding environment.
However, the above RL formulation presupposes two critical conditions for decision policy training:
1) an interactable environment, and
2) a pre-specified reward function.
Regrettably, online interactions with the environment can be both expensive and hazardous~\citep{mihatsch2002risk,hans2008safe,garcia2015comprehensive}, while developing a suitable reward function typically necessitates considerable human effort.
Additionally, the heuristic rewards often employed may be insufficient to express the true intent~\citep{hadfield2017inverse}.

To address these challenges, prior research has explored two approaches.
First, some works have focused on the offline RL formulation~\citep{fujimoto2019off}, where the learner has access to fixed offline trajectories along with a reward signal for each transition (or limited expert demonstrations).
Second, others have considered the (online) preference-based RL formulation, where the task objective is conveyed to the learner through preferences of a human annotator between two trajectories rather than rewards for each transition.
In pursuit of further advancements in this setting, we propose a novel approach that relaxes both of these requirements and advocates for offline preference-based RL (PbRL).

\begin{figure*}[ht] %htbp
	\centering
	% \vspace{-10pt}
	\includegraphics[width=0.85\textwidth]{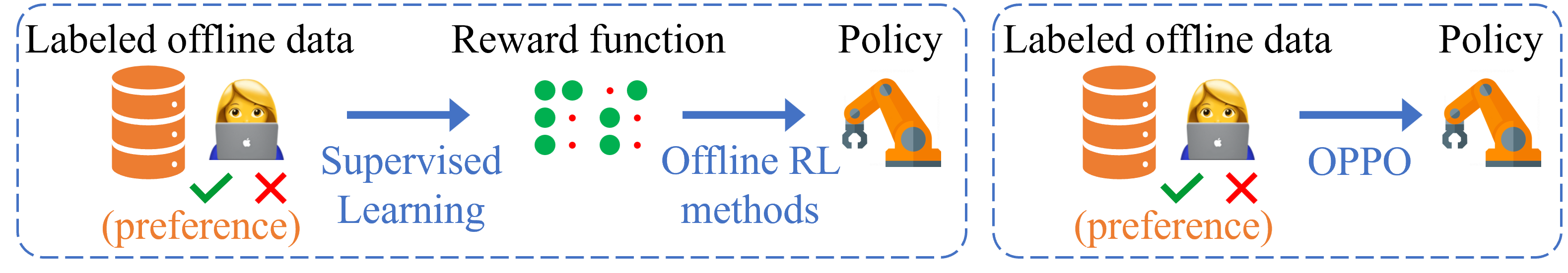}
	\vspace{-10pt}
	\caption{
		A flow diagram of previous offline PbRL algorithms (left) and our OPPO algorithm (right).
		Previous works require learning a separate reward function for modeling human preferences using the Bradley-Terry model.
		In contrast, our OPPO directly optimizes the policy network.
	}
	\label{fig:pbrl-vs-oppo}
	\vspace{-10pt}
\end{figure*}

In the context of offline preference-based reinforcement learning (PbRL), where access to an offline dataset and labeled preferences between the offline trajectories is available, a common approach is to combine previous online PbRL methods with off-the-shelf offline RL algorithms~\citep{shin2021offline}.
This two-step strategy, as illustrated in Fig.\ref{fig:pbrl-vs-oppo} (left), typically involves training a reward function using the Bradley-Terry model~\citep{Bradley1952RankAO} in a supervised manner, followed by training the policy with any offline RL algorithm on the transitions relabeled via the learned reward function.
However, the practice of separately learning a reward function that explains expert preferences may not directly instruct the policy on how to act optimally.
This is because preference labels define the PbRL task, and the goal is to learn the most preferred trajectory by the annotator rather than to maximize the cumulative discounted proxy rewards of the policy rollouts.
In cases of complex tasks, such as non-Markovian tasks, scalar rewards may create an information bottleneck in policy improvement, resulting in suboptimal behavior~\citep{vamplew2022scalar}.
Additionally, an isolated policy optimization may exploit loopholes in miscalibrated reward functions, leading to undesirable behaviors.
Given these limitations, it is reasonable to question the necessity of learning a reward function, especially considering that it may not directly yield the optimal policy.

To achieve this objective, we present the offline preference-guided policy optimization (OPPO) approach, which is a one-step paradigm that simultaneously models offline preferences and learns the optimal decision policy without requiring the separate learning of a reward function (as illustrated in Figure \ref{fig:pbrl-vs-oppo} right).
This is achieved through the use of two objectives: an offline hindsight information matching objective and a preference modeling objective.
By iteratively optimizing these objectives, we derive a contextual policy $\pi(\ba|\bs,\bz)$ to model the offline data and an optimal context $\bz^*$ to model the preference.
The main focus of OPPO is on both learning a high-dimensional $\bz$-space and evaluating policies within such space.
This high-dimensional $\bz$-space captures more task-related information compared to scalar reward, making it ideal for policy optimization purposes.
Furthermore, the optimal policy is obtained by conditioning the contextual policy $\pi(\ba|\bs,\bz)$ on the learned optimal context $\bz^*$.

Our main contribution can be summarized as follows.
Firstly, we propose OPPO, a concise, stable, and one-step offline PbRL paradigm that avoids the need for separate reward function learning.
Secondly, we present an instance of a preference-based hindsight information matching objective and a novel preference modeling objective over the context.
Finally, extensive experiments are conducted to demonstrate the superiority of OPPO over previous competitive baselines and to analyze its performance.

% Unfortunately, many IRL algorithms are extremely expensive to run, requiring reinforcement learning in an inner loop.
% Scaling IRL methods to large environments has thus been the focus of much recent work [7, 14].
% Fundamentally, however, IRL learns a cost function, which explains expert behavior but does not directly tell the learner how to act.
% Given that learner's true goal often is to take actions imitating the expert-indeed, many IRL algorithms are evaluated on the quality of the optimal actions of the costs they learn-why, then, must we learn a cost function, if doing so possibly incurs significant computational expense yet fails to directly yield actions?

\section{Related Work}
Since OPPO is at the intersection of Preference-Based Reinforcement Learning, Offline RL, and conditional RL we review the most relevant algorithms from these fields (see Table \ref{tab:relatedwork})

\begin{table*}[ht]
    \small
    \centering
    \caption{A concise tabular representation of the differences between our method and related works.}
    \label{tab:relatedwork}
    \begin{tabular}{lllllc}
    \toprule
                       & \textbf{Method}    & \textbf{Supervised Signal}    & \textbf{Training} & \textbf{Architectures} & \makecell{\textbf{Learning a Separate} \\ \textbf{Reward Function}} \\ \midrule
    Imitation Learning & BC                 & Expert demonstration          & Offline           & MLP                    & $\times$                                                            \\ \midrule
    Online PbRL        & PEBBLE             & Preference                    & Online            & MLP                    & $\checkmark$                                                        \\
                       & SURF               & Preference                    & Online            & MLP                    & $\checkmark$                                                        \\ \midrule
    Offline RL         & CQL                & Ground Truth Reward           & Offline           & MLP                    &                                                                     \\
                       & DT                 & Ground Truth Reward           & Offline           & Transformer            &                                                                     \\ \midrule
    Offline PbRL       & OPAL               & Preference                    & Offline           & MLP                    & $\checkmark$                                                        \\
                       & PT                 & Preference                    & Offline           & Transformer            & $\checkmark$                                                        \\
                       & OPPO               & Preference                    & Offline           & Transformer            & $\times$                                                            \\ \bottomrule
    \end{tabular}
    \vspace{-10pt}
\end{table*}

\paragraph{Online PbRL.}
Preference-based RL is also known as reinforcement learning from human feedback (RLHF).
Several works have successfully utilized feedback from real humans to train RL agents \citep{DilipArumugam2019DeepRL,PaulFChristiano2017DeepRL,BorjaIbarz2018RewardLF,WBradleyKnox2009InteractivelySA,KiminLee2021PEBBLEFI,GarrettWarnell2017DeepTI}.
\citet{PaulFChristiano2017DeepRL} scaled preference-based reinforcement learning to utilize modern deep learning techniques, and \citet{BorjaIbarz2018RewardLF} improved the efficiency of this method by introducing additional forms of feedback such as demonstrations.
Recently, PEBBLE~\citep{KiminLee2021PEBBLEFI} proposed a feedback-efficient RL algorithm by utilizing off-policy learning and pre-training.
SURF~\citep{park2022surf} used pseudo-labeling to utilize unlabeled segments and proposed a novel data augmentation method called temporal cropping.
All of the above methods require the agent to online interact with the environment, and they are all two-step strategies that require learning a scalar reward function separately.

\paragraph{Offline RL.}
To mitigate the impact of distribution shifts in offline RL, prior algorithms
(a) constrain the action space~\citep{fujimoto2019off,kumar2019stabilizing,siegel2020keep,zhuang2023behavior},
(b) incorporate value pessimism~\citep{fujimoto2019off,kumar2020conservative,liu2022dara}, and
(c) introduce pessimism into learned dynamics models~\citep{kidambi2020morel,yu2020mopo}.
Another line of work explored learning a wide behavior distribution from the offline dataset by learning a task-agnostic set of skills, either with likelihood-based approaches~\citep{ajay2020opal,campos2020explore,pertsch2020accelerating,singh2020parrot} or by maximizing mutual information~\citep{eysenbach2018diversity,lu2020reset,sharma2019dynamics}.
% \paragraph{Transformers for RL.}
% Transformers~\citep{vaswani2017attention} have been applied successfully to many tasks in natural language processing~\citep{devlin2018bert,radford2018improving} and computer vision~\citep{carion2020end,dosovitskiy2020image}.
% However, applying the transformer architecture in the RL setting is still an open challenge.
% \citet{loynd2020working} studied how transformer-based models can improve the performance of sequential decision-making agents.
% They stabilized the architecture using factored observations and an intense hyperparameter tuning procedure, resulting in an improved sample efficiency.
% Previous works have also studied transformers replacing LSTMs in imitation learning.
% For example, \citet{dasari2020transformers} studied one-shot imitation learning, and \citet{abramson2020imitating} combined the modalities of language and image for text-conditioned behavior generation.
% Recent work studied how to replace RL algorithms with transformer-based language models~\citep{janner2021offline,chen2021decision}.
% They modeled the agent as a sequence problem using a supervised prediction loss in the offline RL setting.
% In this way, they could use the vanilla transformer without architectural additions or imposing restrictions on the observations.
% \paragraph{Supervised learning in RL.}
Some prior methods for RL is more similar to static supervised learning, such as Q-learning~\citep{watkins1989learning,mnih2013playing} and behavior cloning.
In these mathods, the resulting agent's performance is positively correlated to the quality of data used for training.
In addition to aforementioned RL methods, \citet{srivastava2019training} and \citet{kumar2019reward} studied "upside-down" reinforcement learning (UDRL), seeking to model behaviors via a supervised loss conditioned on a target return.
\citet{ghosh2019learning, liu2022learn, liu2021unsupervised} extended prior UDRL methods to perform goal reaching by taking the goal state as the condition, and \citet{paster2020planning} further used an LSTM for goal-conditioned online RL settings.
DT~\citep{chen2021decision} and TT~\citep{janner2021offline} solved the problem via sequence modeling, since they believe sequence modeling enables to model behaviors without access to the reward, in a similar style to language~\citep{radford2018improving} and images~\citep{chen2020generative}.
% At the same time, it is also known to scale well~\citep{brown2020language}.
Although the above methods can avoid online interaction between the agent and the environment, they all require ground truth reward or expert demonstrations to specify the task, which often requires a lot of human labor.

\paragraph{Offline PbRL.}
OPAL~\citep{shin2021offline} first tried to solve offline PbRL by simply combining previous (online) PbRL method and off-the-shelf offline RL algorithm.
PT~\citep{kim2023preference} introduced a new preference model based on the weighted sum of non-Markovian rewards and utilized transformer-based architecture to design such model.
Both of these works adopt naive two-step strategy with learning a reward function separately.
To avoid information bottleneck that scalar reward may create~\citep{vamplew2022scalar}, OPPO jointly models offline preferences and learns the optimal decision policy in a one-step paradigm.
And in contrast to both supervised RL and UDRL, the purpose of our method is to search for the optimal solution supervised by a binary preference signal in the offline setting.
Our method is not only working with sub-optimal demonstrations but also revealing optimal behaviors without injecting human priors about the optimal demonstration.

\section{Preliminaries}
% Reinforcement learning (RL) is a framework where an agent interacts with an environment in discrete time.
We consider reinforcement learning (RL) in a Markov decision process (MDP) described by a tuple $(\mathcal{S},\mathcal{A},r,P,p_0, \gamma)$, where $\bs_t \in \mathcal{S}$, $\ba_t \in \mathcal{A}$, and $r_t = r(\bs_t, \ba_t)$ denote the state, action, and reward at timestep $t$, $P(\bs_{t+1}|\bs_t,\ba_t)$ denotes the transition dynamics, $p_0(\bs_0)$ denotes the initial state distribution, and $\gamma \in [0,1)$ denotes the discount factor.
At each timestep $t$, the agent receives a state $\bs_t$ from the environment and chooses an action $\ba_t$ based on the policy $\pi(\ba_t|\bs_t)$.
In the standard RL framework, the environment returns a reward $r_t$, and the agent transits to the next state $\bs_{t+1}$.
The expected return $\mathcal{J}_r(\pi) = \mathbb{E}_{\tau\sim \pi(\tau)}\sum^\infty_{k=0} \gamma^k r(s_{t+k},a_{t+k})$ is defined as the expectation of discounted cumulative rewards, where $ \tau = \left( \bs_0, \ba_0, \bs_1, \ba_1, \dots \right)$, $\bs_0 \sim p_0(\bs_0)$, $\ba_t \sim \pi(\ba_t | \bs_t)$, and $\bs_{t+1} \sim P(\bs_{t+1} | \bs_t, \ba_t)$.
The agent's goal is to learn a policy $\pi$ that maximizes the expected return.

\subsection{Offline Preference-based reinforcement learning}\label{OPbRL}
In this work, we assume a fully offline setting in which the agent is not allowed to conduct online rollouts (over the MDP) during training but is provided with a static fixed dataset.
The static dataset, $\mathcal{D} :=\{\tau^0,\dots,\tau^N\}$, consists of pre-collected trajectories, where each trajectory $\tau^i$ contains a contiguous sequence of states and actions: $\tau^i := \{\bs^i_0, \ba^i_0, \bs^i_1, \dots\}$.
Such an offline setting is more challenging than the standard (online) setting as it removes the ability to explore the environment and collect additional feedback.
Unlike imitation learning, we do not assume that the dataset comes from a single expert policy. % attempting to optimize a specific reward function $r(s_t,a_t)$.
Instead, the dataset $\mathcal{D}$ may contain trajectories collected by sub-optimal or even random behavior policies.

Generally, the standard offline RL assumes the existence of reward information for each state-action pair in $\mathcal{D}$.
However, in the offline Preference-based RL (PbRL) framework, we assume that such reward is not accessible, while the agent can access offline preferences (between some pairs of trajectories $(\tau^i, \tau^j)$) that are labeled by an expert (human) annotator.
Specifically, the annotator gives a feedback indicating which trajectory is preferred, i.e., $y \in \{0,1,0.5\}$, where $0$ indicates $\tau^i \succ \tau^j$ (the event that trajectory $\tau^i$ is preferable to trajectory $\tau^j$), $1$ indicates $\tau^j \succ \tau^i$ ($\tau^j$ is preferable to $\tau^i$), and $0.5$ implies an equally preferable case.
Each feedback is stored in a labeled offline dataset $\mathcal{D}_{\succ}$ as a triple $(\tau^i,\tau^j,y)$.
Given these preferences, the goal of PbRL is to find a policy $\pi(\ba_t|\bs_t)$ that maximizes the expected return $\mathcal{J}_{{r}_\psi}$, under the hypothetical reward function ${r}_\psi(\bs_t,\ba_t)$ consistent with human preferences.
To enable this, previous works learn a reward function ${r}_\psi(\bs_t,\ba_t)$ and use the Bradley-Terry model~\citep{Bradley1952RankAO} to model the human preference, expressed here as a logistic function:
\begin{equation}\label{classifier}
    P[\tau^{i} \succ \tau^{j}]
    % =\frac{\exp \sum_{t} \widehat{r}_{\psi}(\mathbf{s}_{t}^{1}, \mathbf{a}_{t}^{1})}
    % % {\exp \sum_{t} \widehat{r}_{\psi}(\mathbf{s}_{t}^{i}, \mathbf{a}_{t}^{i})
    % % +\exp \sum_{t} \widehat{r}_{\psi}(\mathbf{s}_{t}^{j}, \mathbf{a}_{t}^{j})}
    % {\sum_{i \in\{0,1\}} \exp \sum_{t} \widehat{r}_{\psi}(\mathbf{s}_{t}^{i}, \mathbf{a}_{t}^{i})} ,
    = \text{logistic}(\sum_{t} {r}_{\psi}(\mathbf{s}_{t}^{i}, \mathbf{a}_{t}^{i})
    -\sum_{t} {r}_{\psi}(\mathbf{s}_{t}^{j}, \mathbf{a}_{t}^{j})) ,
\end{equation}
where $(\bs_t^i, \ba_t^i) \sim \tau^i$, $(\bs_t^j, \ba_t^j) \sim \tau^j$.
Intuitively, this can be interpreted as the assumption that the probability of preferring a trajectory depends exponentially on the cumulative reward over the trajectory labeled by an underlying reward function.
% While $\widehat r_\psi$ is not a binary classifier, learning $\widehat r_\psi$ amounts to binary classification with labels $y$ provided by a annotator.
The reward function is then updated by minimizing the following cross-entropy loss:
\begin{equation}\label{eq:loss-ce-pbrl}
    -\mathop{\mathbb{E}}\limits_{(\tau^{i}, \tau^{j}, y) \sim \mathcal{D}_{\succ}}
    \Big[(1-y) \log P[\tau^{i} \succ \tau^{j}]
        + y \log P[\tau^{j} \succ \tau^{i}]\Big].
\end{equation}
With the learned reward function $r_\psi$ used to label each transition in the dataset, we can adopt an off-the-shelf offline RL algorithm to enable the policy learning.

\subsection{Hindsight Information Matching}\label{HIM}
Beyond the typical iterative (offline) RL framework, information matching (IM)~\citep{furuta2021generalized} has been recently studied as an alternative problem specification in (offline) RL.
The objective of IM in RL is to learn a contextual policy $\pi(\ba|\bs, \bz)$ whose trajectory rollouts satisfy the pre-defined desired information statistics value $\bz$:
\begin{equation}\label{eq:hindsight-im}
    \min _\pi \mathop{\mathbb{E}}\limits_{\bz \sim p(\bz) \atop {\tau_\bz} \sim \pi(\bz)}
    \left[\loss\left(\bz,  I({\tau}_\bz)\right)\right] ,
\end{equation}
where $p(\bz)$ is a prior, and $\pi(\bz)$ denotes the trajectory distribution generated by rolling out $\pi(\ba|\bs,\bz)$ in the environment.
$I(\tau)$ is a function capturing the statistical information of a trajectory $\tau$, such as the distribution statistics of state and reward, like mean, variance~\citep{wainwright2008graphical}, and $\loss$ is a loss function.

On the one hand, if we set $p(\bz)$ as a prior distribution, optimizing Eq.\ref{eq:hindsight-im} corresponds to performing unsupervised (online) RL to learn a set of skills~\citep{eysenbach2018diversity, sharma2019dynamics}. %liu2022learn, liu2021unsupervised
On the other hand, if we set $p(\bz)$ as statistical information of a given off-policy trajectory (or state-action) distribution $\mathcal{D}(\tau)$ (or $\mathcal{D}(\bs,\ba)$), Eq.\ref{eq:hindsight-im} corresponds to an objective for hindsight information matching in (offline) RL.
For example, HER~\citep{andrychowicz2017hindsight} and return-conditioned RL (upside-down RL~\citep{srivastava2019training,kumar2019reward,chen2021decision,janner2021offline}) use the above concept of hindsight: specifying any trajectory $\tau$ in the dataset as the hindsight target and setting the information $\bz$ in Eq.\ref{eq:hindsight-im} as $I(\tau)$.
Then, we provide the $I(\cdot)$-driven hindsight information matching (HIM)  objective:
\begin{equation}
    \min _\pi \mathop{\mathbb{E}}\limits_{\tau \sim \mathcal{D}(\tau) \atop {\tau}_{\bz} \sim \pi(\bz)}
    \left[\loss\left(I({\tau}), I(\tau_\bz)\right)\right],
\end{equation}
where $\bz := I(\tau)$.
In HER, we set $I(\tau)$ as the final state in trajectory $\tau$,
and in reward-conditional RL, we set $I(\tau)$ as the return of trajectory $\tau$.
Thus, we can use the hindsight information $\bz:=I(\tau)$ to provide supervision for training the contextual policy $\pi(\ba|\bs, \bz)$.
However, in the offline setting, sampling $\tau_\bz$ from $\pi(\bz)$ is not accessible.
Thus, we must model the environment transition dynamics besides $I(\cdot)$-driven hindsight information modeling.
That is to say, we need to model the trajectory itself, \ie $\min _\pi \mathbb{E}_{\tau \sim \mathcal{D}(\tau), {\tau}_{\bz} \sim \pi(\bz)}\left[\loss\left({\tau}, \tau_\bz\right)\right]$.
Then, we provide the overall offline HIM objective:
\begin{equation}\label{eq:hindsight-offline-im}
    \min _\pi \mathop{\mathbb{E}}\limits_{\tau \sim \mathcal{D}(\tau) \atop {\tau}_{\bz} \sim \pi(\bz)}
    \left[\loss\left(I({\tau}), I(\tau_\bz)\right) + \loss\left({\tau}, \tau_\bz\right)\right].
\end{equation}
To give an intuitive understanding of the above objective, we provide a simple example: considering hindsight $I(\cdot)$ being the return of a trajectory, optimizing $\loss\left(I({\tau}), I(\tau_\bz)\right)$ ensures that the generated $\tau_\bz$ will reach the same return as $\tau = I^{-1}(\bz)$.
However, in the offline setting, we must ensure that the generated $\tau_\bz$ stays in support of the offline data, eliminating the out-of-distribution (OOD) issue. Thus we minimize $\loss\left({\tau}, \tau_\bz\right)$ approximately.
In implementation, directly optimizing $\loss\left({\tau}, \tau_\bz\right)$ is enough to ensure the hindsight information is matched, \eg $\loss\left(I({\tau}), I(\tau_\bz)\right) < \epsilon$.
Here, we explicitly formalize the $\loss\left(I({\tau}), I(\tau_\bz)\right)$ term with particular emphasis on the requisite of hindsight information matching objective and meanwhile highlight the difference, see Section \ref{OPPO}, between the above HIM objective (taking $I(\cdot)$ as a prior) and our OPPO formulation (requiring learning $I_\theta(\cdot)$).

By optimizing Eq.\ref{eq:hindsight-offline-im}, we can obtain a contextual policy $\pi(\ba|\bs, \bz)$.
In the evaluation phase, the optimal policy $\pi(\ba|\bs,\bz^*)$ can be specified by conditioning the policy on a selected target $\bz^*$.
For example, Decision Transformer \citep{chen2021decision} takes the desired performance as the target $\bz^*$(e.g., specify maximum possible return to generate expert behavior),
and RvS-G \citep{emmons2021rvs} takes the goal state as the target $\bz^*$.

\begin{figure*}[t] %htbp
	\centering
	\includegraphics[width=0.85\textwidth]{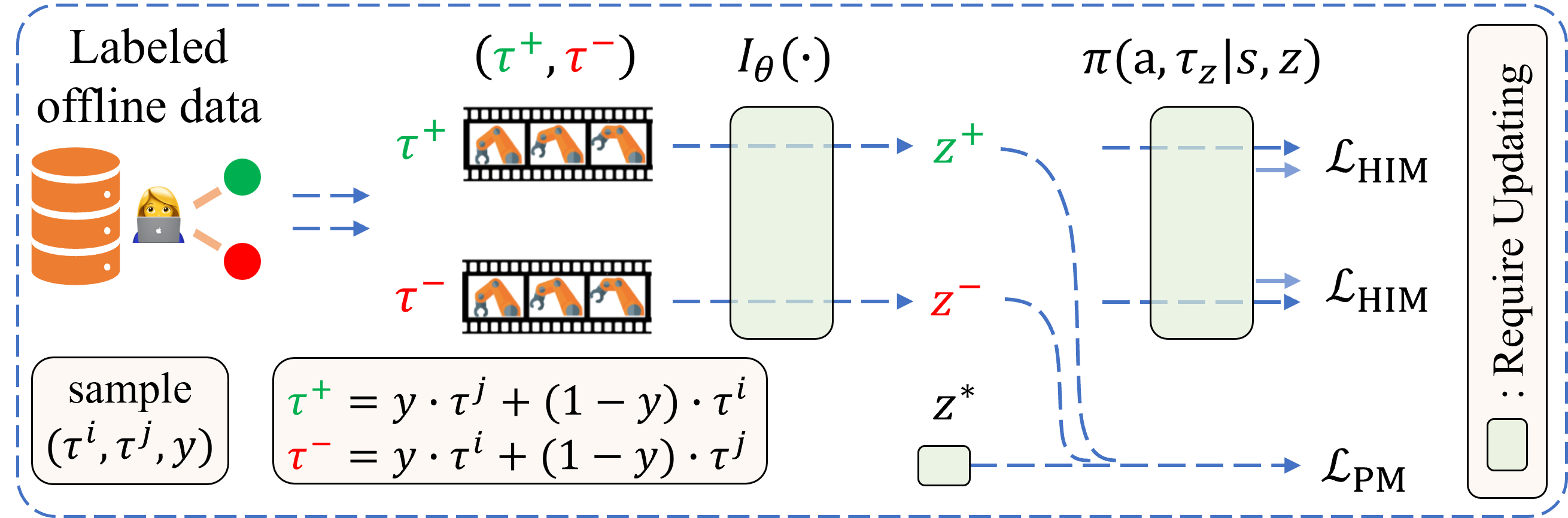}
	% \vspace{-10pt}
	\caption{
		OPPO first maps offline trajectories (both positive $\tau^+$ and negative $\tau^-$) to a latent space via the hindsight information extractor $I_\theta$.
		It then optimizes the offline HIM objective $\mathcal{L}_\text{HIM}$.
		Finally, the belief of the optimal hindsight information $\bz^*$ is updated to model the human preference with objective $\mathcal{L}_\text{PM}$. Meanwhile, the preference modeling loss also regularizes the learning of the hindsight information extractor $I_\theta$.
	}
	\label{fig:oppo-framework}
	\vspace{-5pt}
\end{figure*}

\section{OPPO: Offline Preference-guided Policy Optimization} \label{OPPO}

In this section, we present our method, OPPO (offline preference-guided policy optimization), which adopts the hindsight information matching (HIM) objective in Section~\ref{HIM-obj} to model an offline contextual policy $\pi(\ba|\bs,\bz)$, and introduces a triplet loss in Section~\ref{preference_modeling} to model the human preference as well as the optimal context $\bz^*$.
At testing, we condition the policy on the optimal context $\bz^*$ and thus conduct rollout with $\pi(\ba|\bs,\bz^*)$.
In principle, OPPO is compatible with any PbRL setting, including both online and offline.
In the scope of our analysis and experiments, however, we focus on the offline setting to decouple exploration difficulties in online RL.

\subsection{HIM-driven Policy Optimization}\label{HIM-obj}
As described in Section~\ref{OPbRL}, to directly implement the off-the-shelf offline RL algorithms, previous works in PbRL explicitly learn a reward function with Eq.\ref{eq:loss-ce-pbrl} (as shown in Fig.\ref{fig:pbrl-vs-oppo} left).
As an alternative to such a two-step approach, we seek to learn the policy directly from the preference-labeled offline dataset (as shown in Fig.\ref{fig:pbrl-vs-oppo} right).
Inspired by the offline HIM objective in Section~\ref{HIM}, we propose to learn a contextual policy $\pi(\ba|\bs,\bz)$ in the offline PbRL setting.
Assuming $I_\theta$ being a (learnable) network that encodes the hindsight information in PbRL, we formulate the following objective:
\begin{equation}\label{eq:obj-oppo-1}
	\min _{\pi, I_\theta} \mathcal{L}_{\text{HIM}} :=
	\mathop{\mathbb{E}}\limits_{\tau \sim \mathcal{D}(\tau) \atop {\tau}_{\bz} \sim \pi(\bz)}
	\Big[\loss\left(I_\theta({\tau}), I_\theta(\tau_\bz)\right) + \loss\left({\tau}, \tau_\bz\right)\Big],
\end{equation}
where $\bz := I_\theta({\tau}) $.
Note that Eq.\ref{eq:obj-oppo-1} is a different instantiation of Eq.\ref{eq:hindsight-offline-im} in which we learn the hindsight information extractor $I_\theta(\cdot)$ in the PRBL setting, while previous (offline) RL algorithms normally set $I(\cdot)$ to be a prior~\citep{chen2021decision, emmons2021rvs}.
Such an encoder-decoder structure is now similar to Bi-directional Decision Transformer (BDT) proposed by \citep{furuta2021generalized} for offline imitation learning.
However, since expert demonstrations are missing in the PbRL setting, in Section \ref{preference_modeling}, we propose to use the preference labels in $\mathcal{D}_{\succ}$ to extract hindsight information.

\subsection{Preference Modeling}\label{preference_modeling}
To make the hindsight information $I_\theta({\tau})$ in Eq.\ref{eq:obj-oppo-1} match the preference information in the (labeled) dataset $\mathcal{D}_{\succ}$, we construct the following preference modeling objective inspired by the contrastive loss in metric learning \citep{le2020contrastive}:
\begin{equation}\label{eq:obj-oppo-2}
	\min_{\bz^*, I_\theta}\mathop{\mathbb{E}}\limits_{(\tau^{i}, \tau^{j}, y) \sim \mathcal{D}_{\succ}}
	\Big[\loss(\bz^*,\bz^{+}) - \loss(\bz^*,\bz^{-})\Big],
\end{equation}
where $\bz^{+}$ and $\bz^{-}$ represent the embedding of the preferable (positive) trajectory $I_\theta(y\tau^{j}+(1-y)\tau^{i})$ and that of the less preferable (negative) trajectory $I_\theta(y\tau^{i}+(1-y)\tau^{j})$, respectively.
Closing to the idea of using regret for modeling preference~\citep{knox2022models,chen2022human}, our basic assumption of designing the objective in Eq.\ref{eq:obj-oppo-2} is that humans normally conduct two-level comparisons before giving preferences between two trajectories $(\tau^i, \tau^j)$:
1) separately judging the similarity between trajectory $\tau^i$ and the hypothetical optimal trajectory $\tau^*$, i.e.  $-\loss(\bz^*,\bz^i)$, and the similarity between trajectory $\tau^j$ and the hypothetical optimal one $\tau^*$, $-\loss(\bz^*,\bz^j)$, and
2) judging the difference between the above two similarities ($-\loss(\bz^*,\bz^i)$ vs. $-\loss(\bz^*,\bz^j)$) and setting the trajectory with the higher similarity as the preferred one.
Hence, optimizing Eq.\ref{eq:obj-oppo-2} guarantees finding the optimal embedding that is more similar to $\bz^{+}$ and less similar to $\bz^{-}$.
To clarify, $\bz^*$ is the corresponding  contextual information for $\tau^*$, whereas $\tau^*$ will always be preferred over any offline trajectories in the dataset.

\begin{algorithm}[t]
	% \vspace{-10pt}
	\caption{OPPO: Offline Preference-guided Policy Optimization}
	\label{alg:oppo-training}
	\textbf{Require:} Dataset $\mathcal{D} := \{\tau \} $ and labeled dataset $\mathcal{D}_{\succ} := \{ (\tau^i, \tau^j, y) \}$, where $\tau^i \in \mathcal{D}$ and $\tau^j \in \mathcal{D}$.
	\textbf{Return:} $\pi(\ba|\bs, \bz)$ and $\bz^*$.
	\begin{algorithmic}[1]
		\STATE Initialize policy network $\pi(\ba|\bs, \bz)$, hindsight information extractor $I_\theta: \tau \to \bz$, and the optimal context embedding $\bz^*$.
		\WHILE{not converged}
		\STATE Sample a batch of trajectories from $\mathcal{D}$: $\{\tau\}_{\text{B}} \sim \mathcal{D}$.
		\STATE Update $\pi(\ba|\bs, \bz)$ and $I_\theta(\cdot)$ with sampled $\{\tau\}_{\text{B}}$ using $\mathcal{L}_{\text{HIM}}$. %Eq.\ref{eq:obj-oppo-1}.
		\STATE Sample a batch of preferences from $\mathcal{D}_{\succ}$: $\{(\tau^i, \tau^j, y)\}_{\text{B}} \sim \mathcal{D}_{\succ}$.
		\STATE Update $I_\theta(\cdot)$ and the optimal $\bz^*$ with sampled $\{(\tau^i, \tau^j, y)\}_{\text{B}}$ using $\mathcal{L}_{\text{PM}}$. % Eq.\ref{eq:obj-oppo-3}.
		\ENDWHILE
	\end{algorithmic}
\end{algorithm}
% \vspace{-40pt}

In practice, to robustify the preference modeling, we optimize the following objective using the triplet loss in place of the objective in Eq.\ref{eq:obj-oppo-2}:
\begin{equation}\label{eq:obj-oppo-3}
	\min_{\bz^*, I_\theta} \mathcal{L}_{\text{PM}} := \mathbb{E}%_{(\tau^{i}, \tau^{j}, y) \sim \mathcal{D}_{\succ}}
	\Big[\max(\loss(\bz^*,\bz^{+}) - \loss(\bz^*,\bz^{-})+m, 0 )\Big],
\end{equation}
where $m$ is an arbitrarily set margin between positive and negative pairs.
It is worth mentioning that the posterior of the optimal embedding $\bz^*$ and the hindsight information extractor $I_\theta(\cdot)$ are updated alternatively to ensure learning stability.
A better estimate of the optimal embedding helps the encoder to extract features to which the human labeler pays more attention.
In contrast, a better hindsight information encoder, on the other hand, accelerates the search process for the optimal trajectory in the high-level embedding space.
In this way, the loss function for the encoder consists of two parts:
1) a hindsight information matching loss in a supervised style as in Eq.\ref{eq:obj-oppo-1} and
2) a triplet loss as in Eq.\ref{eq:obj-oppo-3} to better incorporate the binary supervision provided by the preference-labeled dataset.

\subsection{Training Objectives \& Implementation Details}
In our experiment, we consolidate $\loss$ in Eq.\ref{eq:obj-oppo-1} as MSE Loss and in Eq.\ref{eq:obj-oppo-3} as Euclidean Distance.
In this case, we model $\bz^*$ as a point in the $\bz-$space, and the similarity measure $\ell$ is $L_2$ distance.
An alternative option is to model $\bz^*$ as a point sampled from a learned distribution in the $\bz-$space, where $\ell$ is a measurement between two distributions, such as the KL divergence.
Also, we add a normalization loss $\mathcal{L}_{\text{norm}}$ to constrain the L2 norm of all embeddings generated by hindsight information extractor $I_\theta$.
\begin{equation}\label{eq:total-loss}
    \mathcal{L}_{\text{total}} := \mathcal{L}_{\text{HIM}} + \alpha \mathcal{L}_{\text{PM}} + \beta \mathcal{L}_{\text{norm}}
\end{equation}
% \paragraph{Architecture \& Implementation Details} \label{implementation}
The architecture overview of OPPO is shown in Fig.\ref{fig:oppo-framework}.
OPPO models the hindsight information extractor $I_{\theta}$ as an encoder network $I_\theta: \tau \to \bz$ and we use the BERT architecture.
Furthermore, similar to DT~\citep{chen2021decision}, we use the GPT architecture to model $\pi(\ba |\bs, \bz)$, which predicts future actions via autoregressive modeling.
For specific hyperparameter selection during the training process, please refer to the detailed description in Appendix \ref{hyperparameters}.
Algorithm~\ref{alg:oppo-training} details the training of OPPO, and the entire process is summarized as follows:
1) We sample a batch of trajectories from the dataset $\mathcal{D}$;
2) In Line 4, we use Eq.\ref{eq:obj-oppo-1} (the hindsight information matching loss) to update $\pi(\ba|\bs, \bz)$ and $I_\theta(\cdot)$ based on sampled trajectories;
consequently, given the $\bz$ extracted out of an offline trajectory by the extractor, the policy is able to reconstruct it;
3) Then, we sample a batch of preferences from the labeled dataset $\mathcal{D}_{\succ}$;
4) Finally, in Line 6, we update $I_\theta(\cdot)$ and $\bz^*$ based on the sampled $\{(\tau^i, \tau^j, y)\}_{\text{B}}$ using Eq.\ref{eq:obj-oppo-3}, making the optimal embedding $\bz^*$ near to the more preferred trajectory $\bz^+$, and meanwhile further away from the less preferred trajectory $\bz^-$.
% ~$\pi(\tau_{\bz^*})$.

\textbf{In summary}, OPPO learns a contextual policy $\pi(\ba|\bs,\bz)$, a context (hindsight information) encoder $I_\theta(\tau)$, and the optimal context, $\bz^*$, for the optimal trajectory $\tau^*$.
Compared with previous PbRL works (first learning a reward function with Eq.\ref{eq:loss-ce-pbrl} and then learning offline policy with off-the-shelf offline RL algorithms), OPPO learns the optimal (offline) policy ($\pi(\ba|\bs,\bz*)$) directly and thus avoids the potential information bottleneck caused by the limited information capacity of scalar reward assignment.
Compared with the HIM-based offline RL algorithms (\eg DT~\citep{chen2021decision} and RvS-G~\citep{emmons2021rvs}), OPPO does not need to manually specify the target context for the rollout policy $\pi(\ba|\bs,\cdot)$ at the testing phase.

% and models the contextual policy with two heads $\pi(\ba, \tau_\bz|\bs, \bz)$, where one head represents outputting (GIVE A LOOK) the current decision $\ba$ and the other head represents modeling the entire trajectory (indicated by Eq.\ref{eq:obj-oppo-1}).
% In implementation, similar to \citet{chen2021decision}, we use the transformer network to model the entire trajectory $\tau$ and directly set the last action in the modeled trajectory as the predicted action $\ba$.

% \begin{equation}\label{dist_classifier}
%     P[\sigma^{1} \succ \sigma^{0}]
%     % =\frac{\exp f(\sigma^1, \sigma^*)}
%     % {\sum_{i \in\{0,1\}} \exp f(\sigma^i, \sigma^*)} ,
% 	= logistic( f(\sigma^1, \sigma^*) - f(\sigma^0, \sigma^*)) ,
% \end{equation}
% However, the search in $\sigma$ space is obviously difficult, so we use the autoencoder structure to train to obtain the hidden space of $\sigma$ and search for the embedding of the optimal trajectory $z^*$ on the hidden space.
% \begin{equation}\label{z_classifier}
%     P_{z^*}[\sigma^{1} \succ \sigma^{0}]
%     % =\frac{\exp f(\sigma^1, \sigma^*)}
%     % {\sum_{i \in\{0,1\}} \exp f(\sigma^i, \sigma^*)} ,
% 	= logistic( f(E(z^1|\sigma^1), z^*) - f(E(z^0|\sigma^0), z^*)) ,
% \end{equation}

\begin{figure*}[ht]
	\centering
% 	% \includegraphics[width=0.6\textwidth]{exp_fig/baseline/legend.pdf} \\
	% \vspace{-5pt}
    \includegraphics[width=0.28\textwidth]{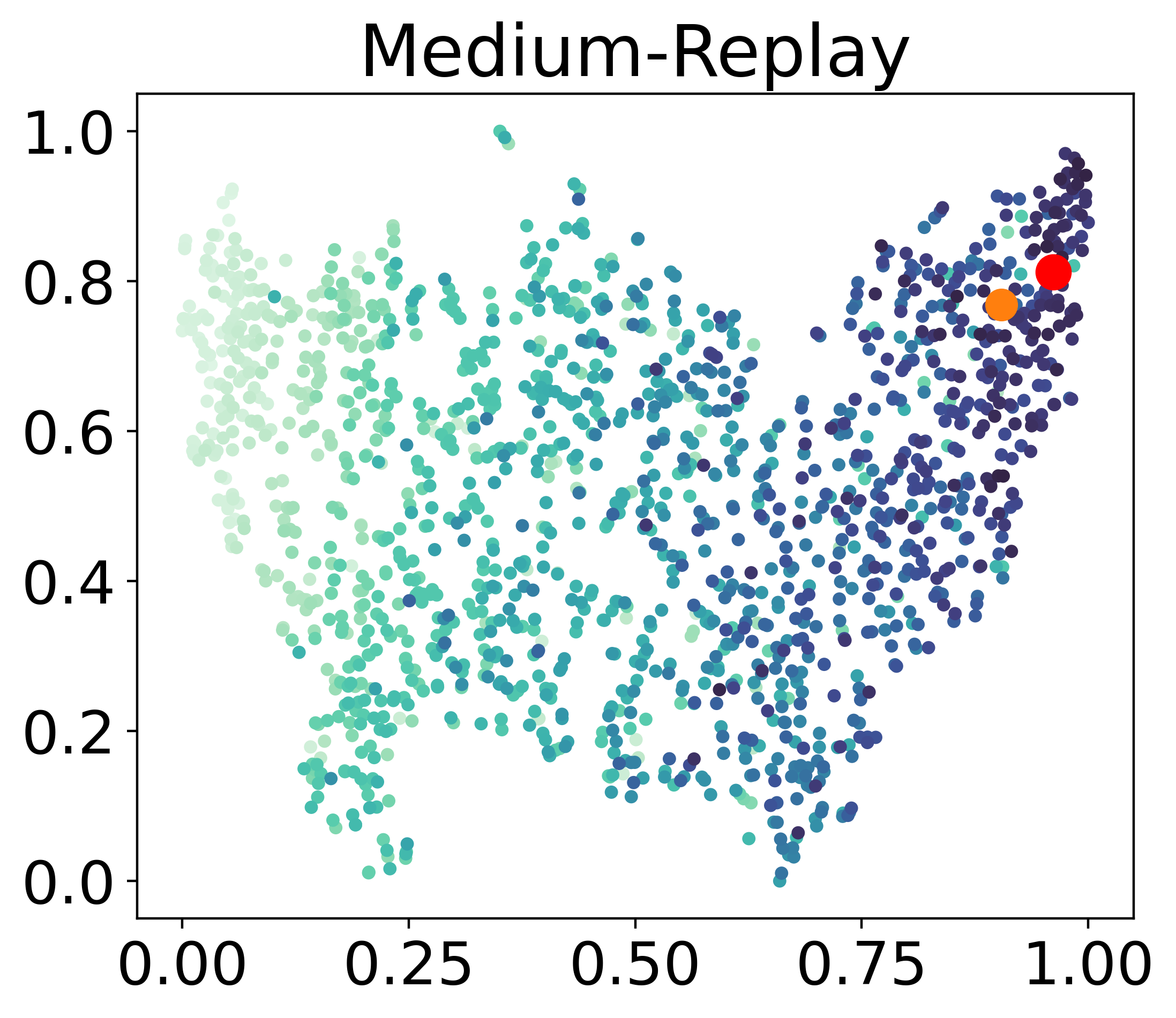}
    \includegraphics[width=0.28\textwidth]{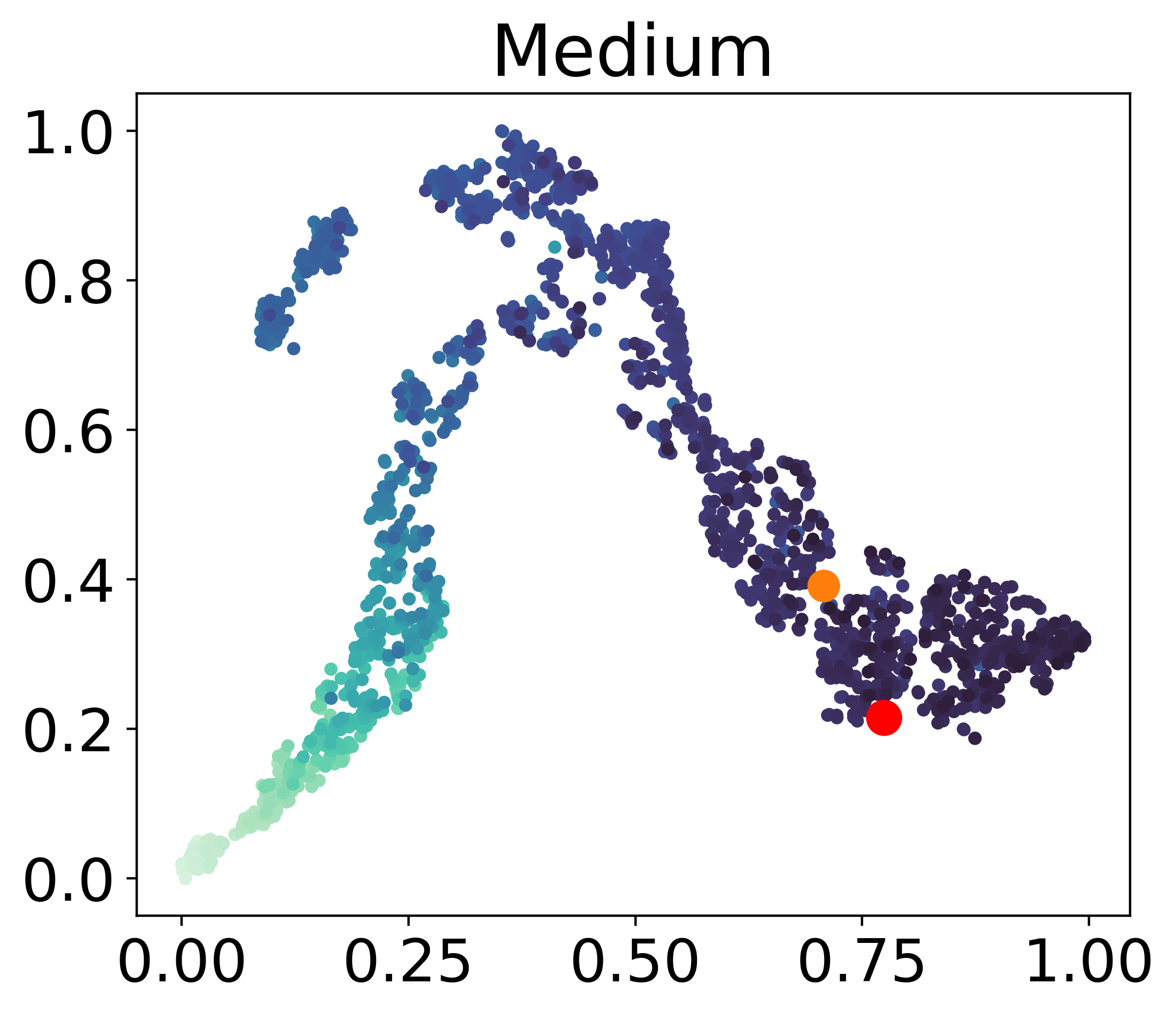}
    \includegraphics[width=0.28\textwidth]{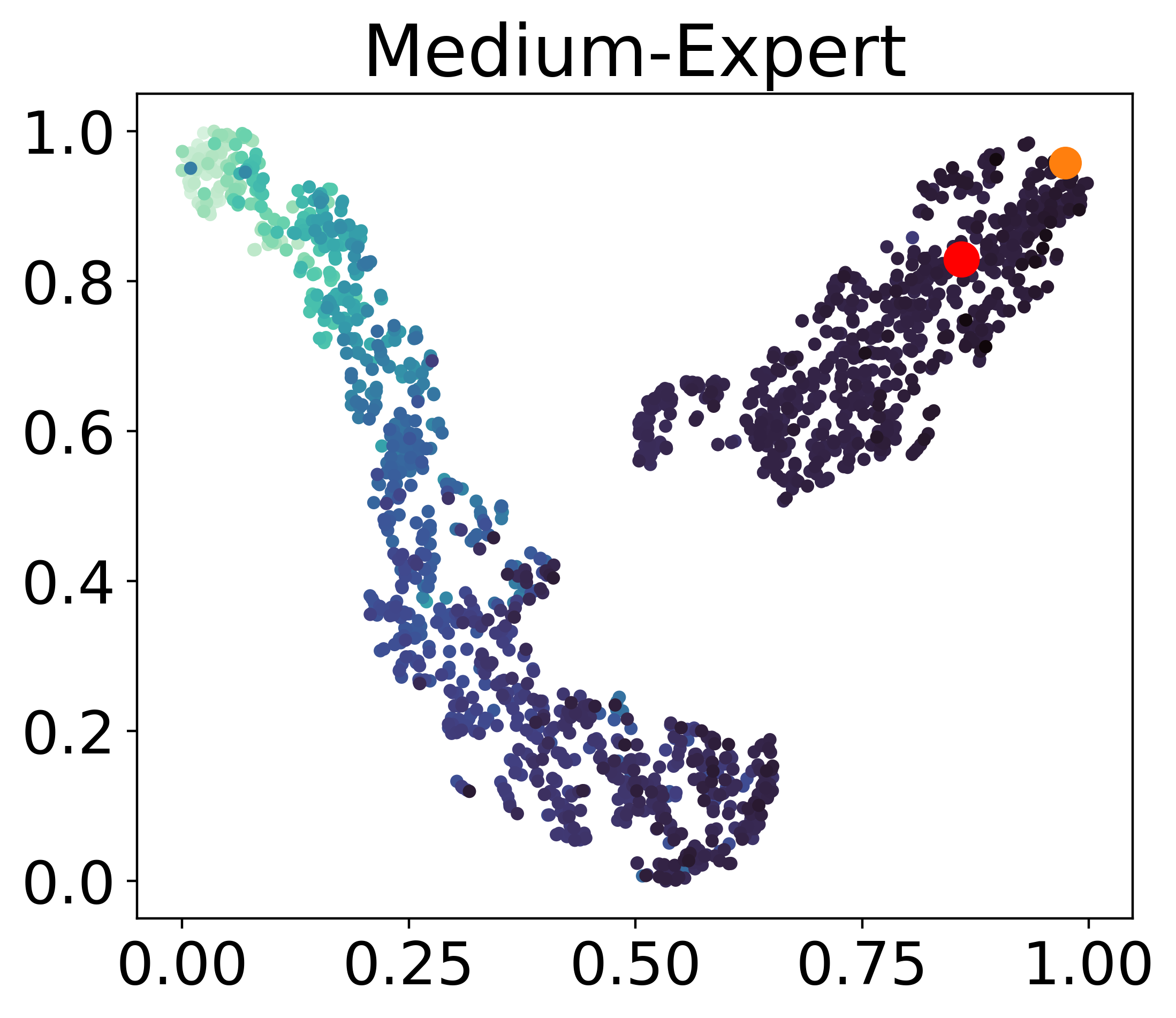}
    \includegraphics[width=0.068\textwidth]{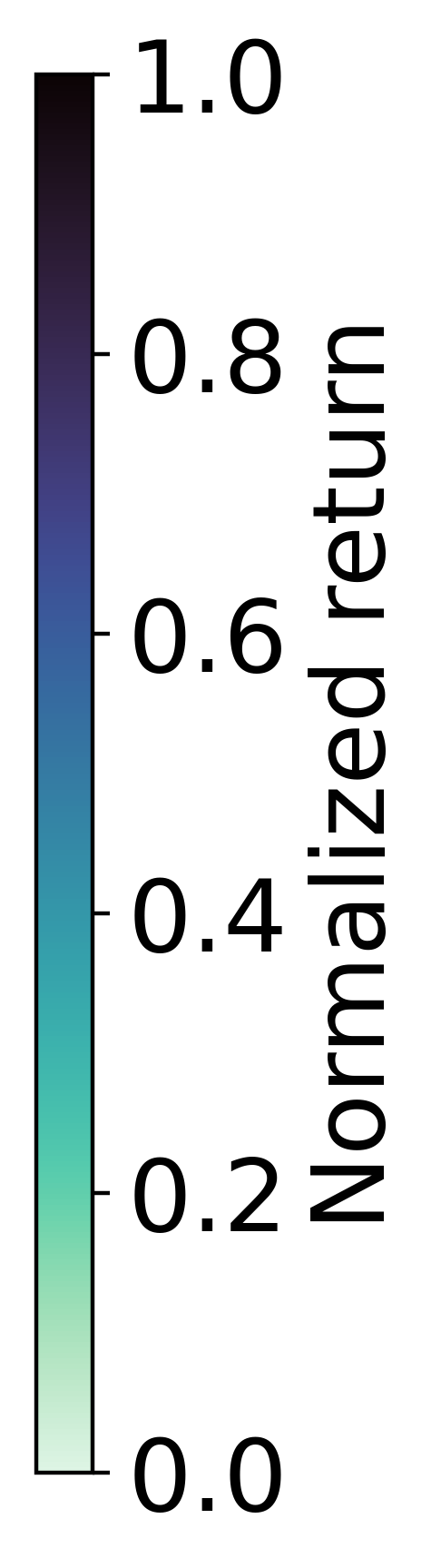}
	% \subfigure[walker]{         \includegraphics[width=0.29\textwidth]{figure/tsne/walker2d_medium-replay.png}
	%                             \includegraphics[width=0.29\textwidth]{figure/tsne/walker2d_medium.png}
	%                             \includegraphics[width=0.29\textwidth]{figure/tsne/walker2d_medium-expert.png}
	%                             \includegraphics[width=0.07\textwidth]{figure/tsne/walker2d_heatmap_legend.png}     } \\
	% \subfigure[halfcheetah]{    \includegraphics[width=0.29\textwidth]{figure/tsne/halfcheetah_medium-replay.png}
	%                             \includegraphics[width=0.29\textwidth]{figure/tsne/halfcheetah_medium.png}
	%                             \includegraphics[width=0.29\textwidth]{figure/tsne/halfcheetah_medium-expert.png}
	%                             \includegraphics[width=0.07\textwidth]{figure/tsne/halfcheetah_heatmap_legend.png}}
	% \vspace{-10pt}
	\caption{
        We utilize t-SNE to visualize the $\bz$-space learned in Hopper environment, encoded with a well-trained $I_\theta(\cdot)$, including the embedding of random trajectories in $\mathcal{D}$, our learned $\bz^*$ (``orange dot") and the actual optimal $\bz^{**}$ (``red dot"), embedding of the best trajectory/policy learned with online reinforcement learning method.
        Color of the points represent the normalized return of the corresponding trajectory $\tau$.
	}
	\label{fig:tsne-q1}
	% \vspace{-5pt}
\end{figure*}

\section{Experiments}

In this section, we evaluate and compare OPPO to other baselines in the offline PbRL setting. A central premise behind the design of OPPO is that the learned hindsight information encoder $I_\theta(\cdot)$ can capture preferences over different trajectories, as described by Eq.\ref{eq:obj-oppo-3}.
Our experiments are therefore designed to answer the following questions:
\begin{enumerate}
\item Does OPPO truly capture the preference? In other words, does the learned $\bz$-space (encoded by the learned $I_\theta(\cdot)$) align with the given preference?
Please refer to Section \ref{sec:zspace}.
\item Can the learned optimal contextual policy $\pi(\ba|\bs, \bz^*)$ outperform the policy $\pi(\ba|\bs, \bz)$ that is conditioned on any other context $\bz \in \{I_\theta(\tau)| \tau \in \mathcal{D} \}$?
Please refer to Section \ref{sec:zpolicy}.
\item Can OPPO achieve the competitive performance compared with other offline PbRL baselines?
Please refer to Section \ref{sec:oppo-performance}.
\item What benefits can we gain from designing the one-step offline PbRL, \ie iteratively conducting offline data modeling (Eq.\ref{eq:obj-oppo-1}) and preference modeling (Eq.\ref{eq:obj-oppo-3})?
Please refer to Section \ref{sec:one-step}.
\item How does OPPO behave in terms of the amount of preference feedback?
Please refer to Section \ref{sec:diff-amount feedback}.
\item Can OPPO attain satisfactory results by incorporating preference from real human instead of scripted teacher?
Please refer to Section \ref{sec:real_human_teacher}.
\end{enumerate}

To answer the above questions, we evaluate OPPO on the continuous control tasks from the D4RL benchmark~\citep{fu2020d4rl}.
Specifically, we choose Hopper, Walker, and Halfcheetah as three base tasks, with medium, medium-replay, medium-expert as the datasets for each task.

\begin{table}[ht]
        \centering
        % \vspace{-10pt}
        \caption{
            Comparison of (normalized) performance when rollouting the contextual policy $\pi(\ba|\bs, \cdot)$ conditioned on different context ($\bz^*$, $\bz_\text{high}$, and $\bz_\text{low}$).
            }
        \label{tab:context-policy-z-q2}
        \resizebox{0.48\textwidth}{!}{
        \begin{tabular}{llrrrrrrrr}
            \toprule
            \multicolumn{1}{c}{\textbf{Environment}} & \multicolumn{1}{c}{\textbf{Dataset}} & \multicolumn{1}{c}{$\bz^*$}    & \multicolumn{1}{c}{$\bz_\text{high}$} & \multicolumn{1}{c}{$\bz_\text{low}$} \\
            \midrule
            \multirow{3}[0]{*}{Hopper}               & Medium-Expert                        & \textbf{108.0 $\pm$ 5.1}       & 94.2  $\pm$ 24.3                      & 79.1 $\pm$ 28.8                      \\
                                                     & Medium                               & \textbf{86.3  $\pm$ 3.2}       & 55.8  $\pm$ 7.9                       & 51.6 $\pm$ 13.8                      \\
                                                     & Medium-Replay                        & \textbf{88.9  $\pm$ 2.3}       & 78.6  $\pm$ 26.3                      & 26.6 $\pm$ 15.2                      \\ \midrule
            \multirow{3}[0]{*}{Walker}               & Medium-Expert                        & \textbf{105.0 $\pm$ 2.4}       & \textbf{106.5 $\pm$ 9.1}              & 93.4 $\pm$ 7.4                       \\
                                                     & Medium                               & \textbf{85.0  $\pm$ 2.9}       & 64.9  $\pm$ 24.9                      & 72.6 $\pm$ 10.6                      \\
                                                     & Medium-Replay                        & \textbf{71.7  $\pm$ 4.4}       & 55.7  $\pm$ 24.8                      & 6.8  $\pm$ 1.7                       \\ \midrule
            \multirow{3}[0]{*}{Halfcheetah}          & Medium-Expert                        & \textbf{89.6  $\pm$ 0.8}       & 48.3  $\pm$ 14.4                      & 42.6 $\pm$ 2.6                       \\
                                                     & Medium                               & \textbf{43.4  $\pm$ 0.2}       & 42.5  $\pm$ 3.9                       & 42.4 $\pm$ 3.2                       \\
                                                     & Medium-Replay                        & \textbf{39.8  $\pm$ 0.2}       & 35.6  $\pm$ 8.5                       & 33.9 $\pm$ 9.2                       \\ \midrule
            \multicolumn{2}{c}{\textbf{Sum}}                                                & \textbf{717.7}                 & 581.9                                 & 448.9                                \\ \bottomrule
            \end{tabular}}
        % \vspace{-10pt}
\end{table}

\subsection{Can $\bz$-space align well with given preferences?}\label{sec:zspace}

In this subsection, we probe that OPPO can enable well-aligned preferences over the $\bz$-space encoded by the learned $I_\theta$.
We first sample random trajectories from the offline dataset $\mathcal{D}$, and encode them with the learned $I_\theta$, and utilize t-SNE~\citep{tsne} as a tool to visualize the encoded $\bz$, shown in Fig.\ref{fig:tsne-q1}.
The learned optimal $\bz^*$ is marked with an orange dot.
Besides, we also mark the (embedding of) optimal trajectory in the D4RL expert dataset, generated by the learned online optimal policy, with a red dot ($\bz^{**}$).

According to Eq.\ref{eq:obj-oppo-3}, embeddings near the actual optimal $\bz^{**}$ in $\bz$-space means they are more preferable implied by the preference label.
Comparing the sampled trajectories (embeddings), we find OPPO successfully captures the preference.
As illustrated in Fig.\ref{fig:tsne-q1}, the trajectories (embeddings) that are near $\bz^{**}$ often have high returns (points with a deeper color).
Further, we observe that our learned optimal $\bz^*$ constantly stays close to actual optimal $\bz^{**}$, which suggests that our learned $\bz^*$ preserves near-optimal behaviors.
Thus, it gives justification that OPPO can make meaningful preference modeling.
%In this subsection, we utilize t-SNE as a tool to visualize the learned $\bz$-space which turns out to be well aligned with the total rewards of these sub-trajectories, as shown in Fig.\ref{tsne_of_pdt}. By analyzing the t-SNE results, we observe that most $\bz^*$ are located in the region of higher normalized return (points with a deeper color). As for walker medium and hopper medium-replay, the nature of uniformness, as we believe, poses difficulties over learning $\bz^*$.
%This shows our method, PDT, has the ability to learn a meaningful $\bz^*$, which is more similar to sub-trajectories with higher normalized return, than to those with lower normalized return.

\subsection{Can $\pi(\ba|\bs, \bz^*)$ achieve better performance?}\label{sec:zpolicy}
Fig.\ref{fig:tsne-q1} shows that our learned $I_\theta(\cdot)$ can produce a well-aligned context embedding $\bz$-space exhibiting effective preference modeling across (embeddings of) trajectories.
More importantly, context embeddings' preference property should be preserved when we condition the context on the learned contextual policy $\pi(\ba|\bs, \cdot)$.
In other words, $I_\theta(\cdot)$ should transfers the preference relationship from $(\tau^i, \tau^j)$ to $(\loss(\bz^i, \bz^*), \loss(\bz^j, \bz^*))$;
further, rolling out the contextual policy $\pi(\ba|\bs,\cdot)$, $(\tau_{\bz^i}, \tau_{\bz^j})$ should similarly preserve the above preference relationship.

\begin{table*}[th]
    \small
    \centering
    % \vspace{-10pt}
    \caption{
        Performance comparison between OPPO and 4 offline (PbRL) baselines (DT+$r$, DT+$r_\psi$, CQL+$r$, and IQL+$r$) in D4RL Gym-Mujoco tasks, where results are reported over 3 seeds.
        }
    \label{tab:results-q3}
    \begin{tabular}{llrrrrrr}
    \toprule
    \multicolumn{1}{c}{\textbf{Environment}} & \multicolumn{1}{c}{\textbf{Dataset}} & \multicolumn{1}{c}{\textbf{Ours}}   & \multicolumn{1}{c}{\textbf{DT+$r$}}  & \multicolumn{1}{c}{\textbf{DT+$r_\psi$}}    & \multicolumn{1}{c}{\textbf{CQL+$r$}} & \multicolumn{1}{c}{\textbf{IQL+$r$}}  & \multicolumn{1}{c}{\textbf{BC}}   \\ \midrule
    \multirow{3}[0]{*}{Hopper}               & Medium-Expert                        & \textbf{108.0 $\pm$ 5.1}            & \textbf{111.0 $\pm$ 0.5}             & 95.6          $\pm$ 27.3                    & \textbf{111.0}                       & 91.5                                  & 79.6                              \\
                                             & Medium                               & \textbf{86.3  $\pm$ 3.2}            & 76.6 $\pm$ 3.9                       & 73.3          $\pm$ 3.0                     & 58.0                                 & 66.3                                  & 63.9                              \\
                                             & Medium-Replay                        & \textbf{88.9  $\pm$ 2.3}            & \textbf{87.8 $\pm$ 4.7}              & 72.5          $\pm$ 22.2                    & 48.6                                 & \textbf{94.7}                         & 27.6                              \\ \midrule
    \multirow{3}[0]{*}{Walker}               & Medium-Expert                        & 105.0 $\pm$ 2.4                     & \textbf{109.2 $\pm$ 0.3}             & \textbf{109.7 $\pm$ 0.1}                    & 98.7                                 & \textbf{109.6}                        & 36.6                              \\
                                             & Medium                               & \textbf{85.0  $\pm$ 2.9}            & 80.9 $\pm$ 3.1                       & 81.1          $\pm$ 2.1                     & 79.2                                 & 78.3                                  & 77.3                              \\
                                             & Medium-Replay                        & 71.7  $\pm$ 4.4                     & \textbf{79.6 $\pm$ 3.1 }             & \textbf{80.4  $\pm$ 4.4}                    & 26.7                                 & 73.9                                  & 36.9                              \\ \midrule
    \multirow{3}[0]{*}{HalfCheetah}          & Medium-Expert                        & \textbf{89.6  $\pm$ 0.8}            & 86.8 $\pm$ 1.3                       & \textbf{88.4  $\pm$ 0.7}                    & 62.4                                 & 86.7                                  & 59.9                              \\
                                             & Medium                               & 43.4  $\pm$ 0.2                     & 43.4 $\pm$ 0.1                       & 43.2          $\pm$ 0.2                     & 44.4                                 & \textbf{47.4}                         & 43.1                              \\
                                             & Medium-Replay                        & 39.8  $\pm$ 0.2                     & 39.2 $\pm$ 0.3                       & 38.8          $\pm$ 0.3                     & \textbf{46.2}                        & 44.2                                  & 4.3                               \\ \midrule
    \multicolumn{2}{c}{\textbf{Sum}}                                                & \textbf{717.7}                      & 714.5                                & 683.0                                       & 575.2                                & 692.4                                 & 429.2                             \\ \bottomrule
    \end{tabular}
    % \vspace{-10pt}
\end{table*}

To show that, we compare the performance of rollouts by the contextual policy $\pi(\ba|\bs, \cdot)$ conditioned on different contexts in Table~\ref{tab:context-policy-z-q2}.
We choose three context embeddings: $\bz^*$, $\bz_\text{high}$ (embedding of the trajectory with the highest return in $\mathcal{D}$), and $\bz_\text{low}$ (embedding of the trajectory with the lowest return in $\mathcal{D}$) and provide respective rollout performances (averaged over 3 seeds).
We discover that the contextual policy $\pi(\ba|\bs, \bz)$ conditioned on $\bz$ with a high (or low) return (of corresponding trajectory $\tau = {I}_\theta^{-1}(\bz)$) obtains an actual high (or low) return when rollouting this policy in the environment, \eg  $\pi(\ba|\bs, \bz_\text{high})$ performs better than $\pi(\ba|\bs, \bz_\text{low})$ (thus preserving the hindsight preference relationship).
Further, when conditioned on the learned optimal $\bz^*$, $\pi(\ba|\bs, \bz^*)$ produces the best performance over that conditioned on all other offline embeddings.
Notice that our learned optimal $\pi(\ba|\bs, \bz^*)$ performs better than the contextual policy  $\pi(\ba|\bs, \bz_\text{high})$.
This result implies that the trajectory of our optimal policy is generally better than other trajectories in the offline dataset.

\subsection{Performance of OPPO on Benchmark Tasks with Scripted Teacher}\label{sec:oppo-performance}
We have shown that OPPO produces a near-optimal context $\bz^*$, and the learned contextual policy $\pi(\ba|\bs, \cdot)$ can preserve the hindsight preference.
This subsection investigates whether the optimal policy $\pi(\ba|\bs, \bz^*)$ can achieve competitive performance on the offline (PBRL) benchmark.
For comparison, we consider three offline PbRL methods and BC as baselines:
1) DT+$r$: performing Decision Transformer~\citep{chen2021decision} with ground-truth reward function, and the results are run by us;
2) DT+$r_\psi$: performing Decision Transformer with a learned reward function (using Eq.\ref{eq:loss-ce-pbrl});
3) CQL+$r$: performing CQL~\citep{kumar2020conservative} with ground-truth reward function;
3) IQL+$r$: performing IQL with ground-truth reward function, the results are reported from IQL~\citep{kostrikov2022offline};
4) BC: performing bahavior cloning on the dataset, the results are reported from DT~\citep{chen2021decision}.

In Table~\ref{tab:results-q3}, we show the performance of OPPO and baselines.
We have the following observations.
1) OPPO has retained a comparable performance against the Decision Transformer trained using true rewards.
% As true reward provides substantially more supervision, the performance of DT+$r$ can be seen as the upper bound for OPPO.
OPPO is a PbRL approach requiring only (human) preferences, which have a more flexible and straightforward form of supervision in the real world.
2) Although DT+$r_\psi$ also shows competitive results in these benchmarks, such a method needs a target of return-to-go determined by the human prior
\footnote{
Preference-based relabelled rewards only participate in the training phase.
During the evaluation phase of DT+$r_\psi$, we pass in the same target return-to-go value as in the original DT paper.
}.
Our method, in contrast, avoids the need for such a prior target by searching across the $\bz$-space.
We argue that our searching method brings advantages because rewards are usually hard to obtain in real-world RL applications, where the preference is the only information accessible for training and deploying an RL method.

% \textcolor{red}{TODO TODO}
% \begin{enumerate}
%     \item Our method, OPPO, has comparable performance against Decision Transformer, which is almost an upper bound for these settings. But our method could work directly under a PbRL setting, i.e. needing only (human) preference instead of true reward, which may be hard to obtain in some real applications.
%     \item Although a conventional PbRL Rewarder, followed by a normal Decision Transformer, also shows competitive result in these benchmarks, these methods actually work under a given true return-to-go value as required by DT\footnote{Preference-based relabelled rewards only participate in training phase. During the evaluation phase of DT(wPr), we pass in the same return-to-go value as of original DT paper}. However, our method, by searching across z-space, avoids the need of such pre-defined hyperparameter. We think this searching method brings advantages as real-world RL application usuall no rewards, and return-to-go could be obtained and preference is the information that could be used to train and \textbf{deploy} a RL method.
% \end{enumerate}

\begin{figure*}[ht]
	\centering
	% \vspace{-5pt}
    \includegraphics[width=0.28\textwidth]{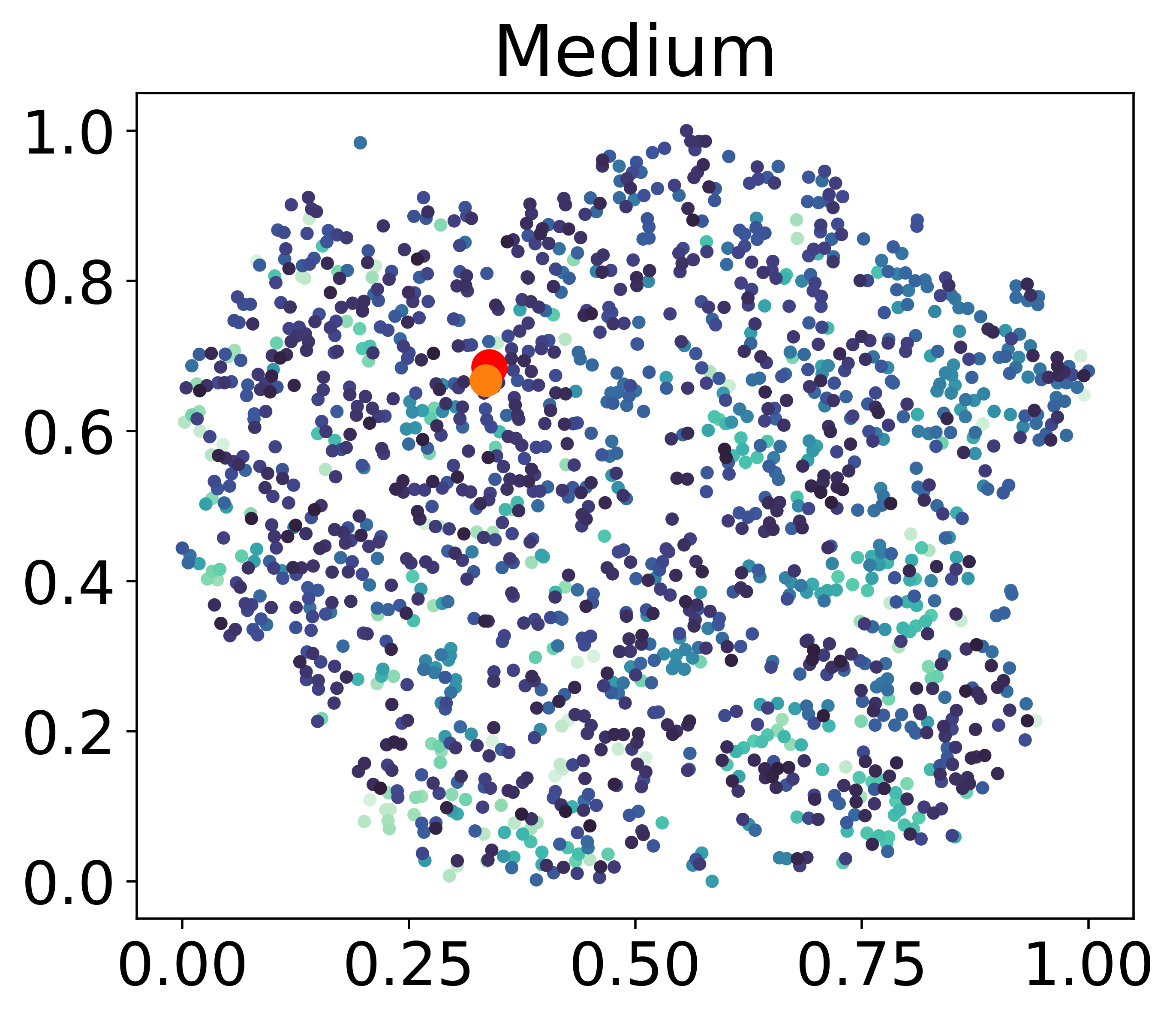}
    \includegraphics[width=0.28\textwidth]{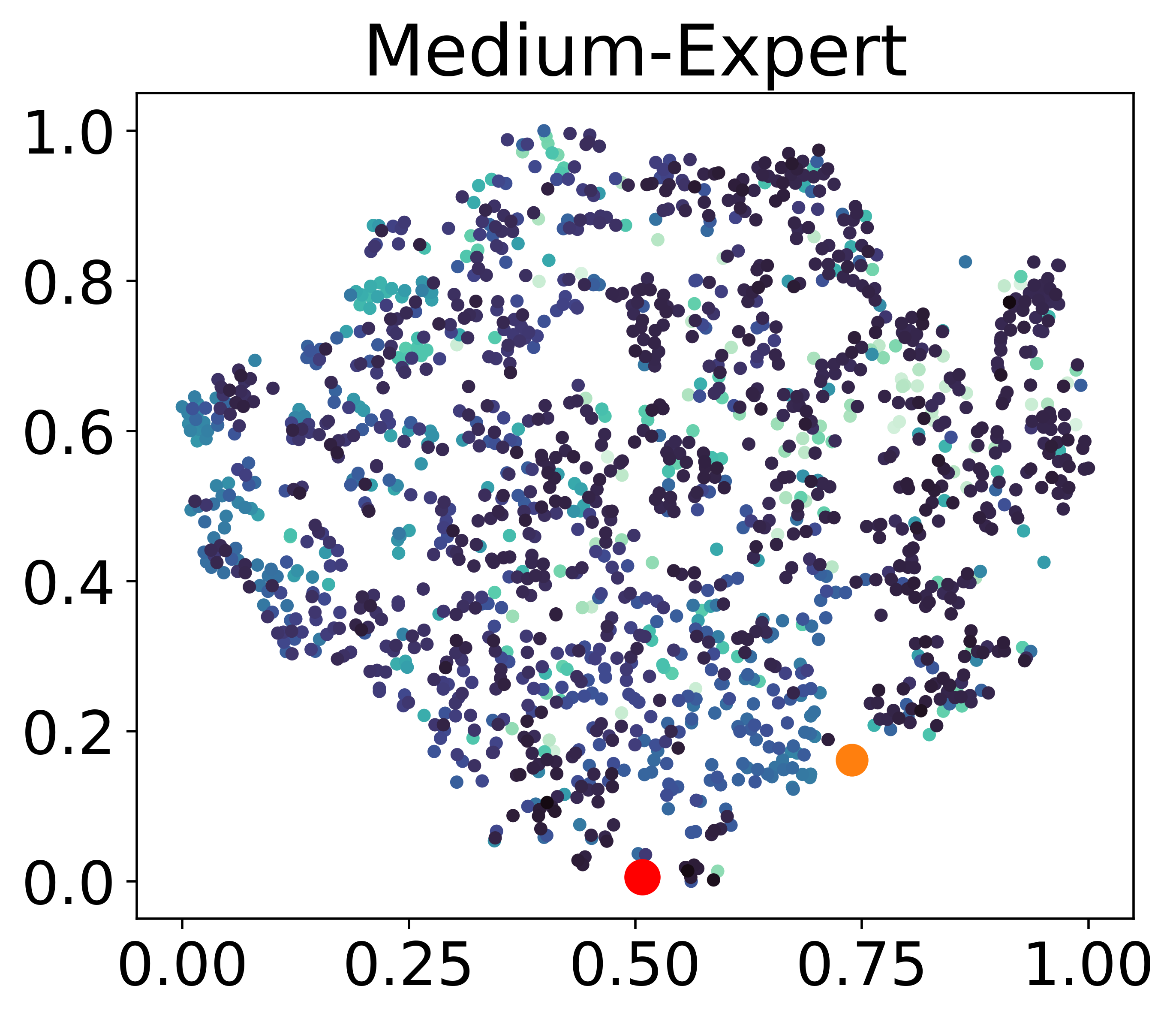}
    \includegraphics[width=0.28\textwidth]{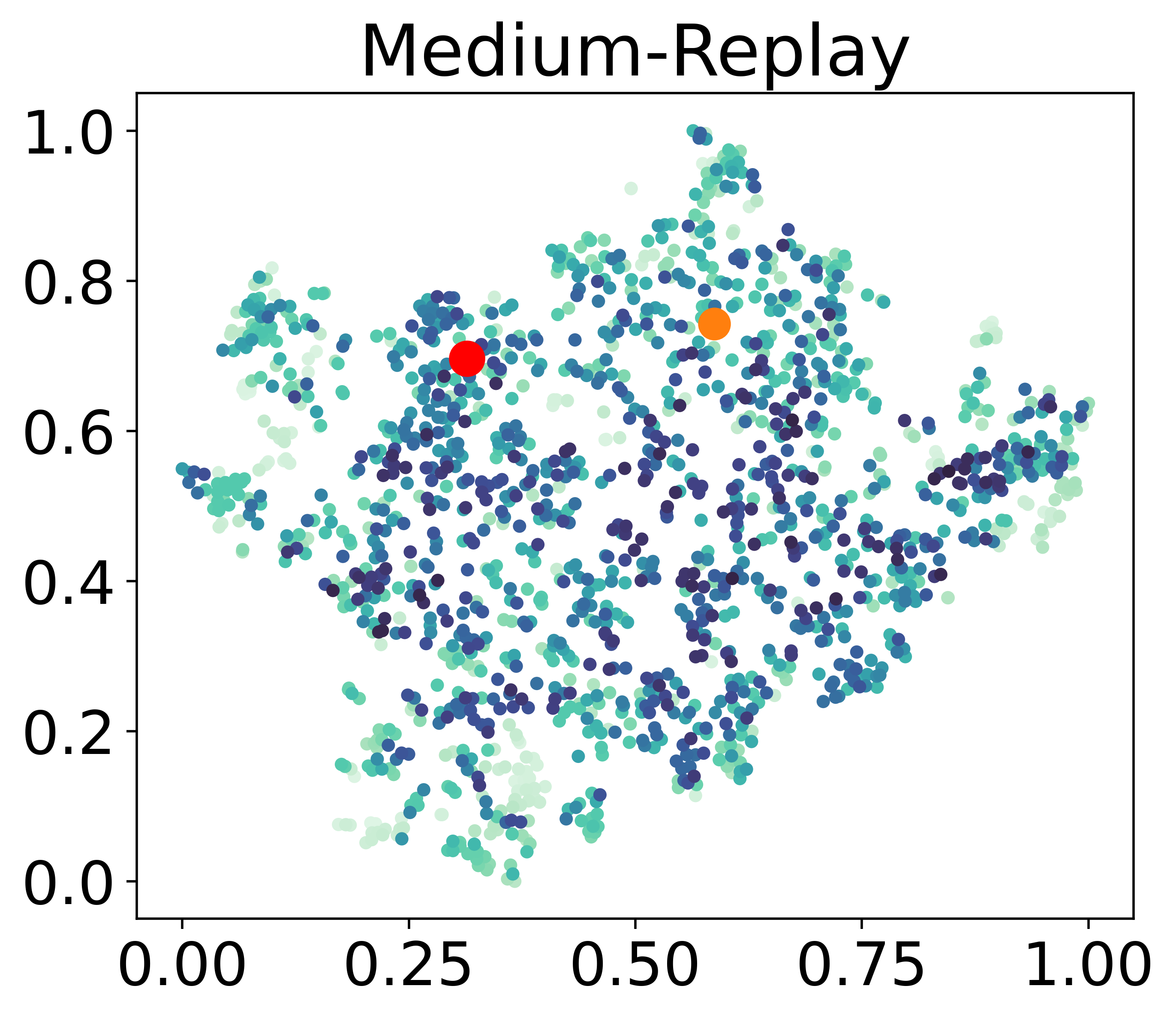}
    \includegraphics[width=0.068\textwidth]{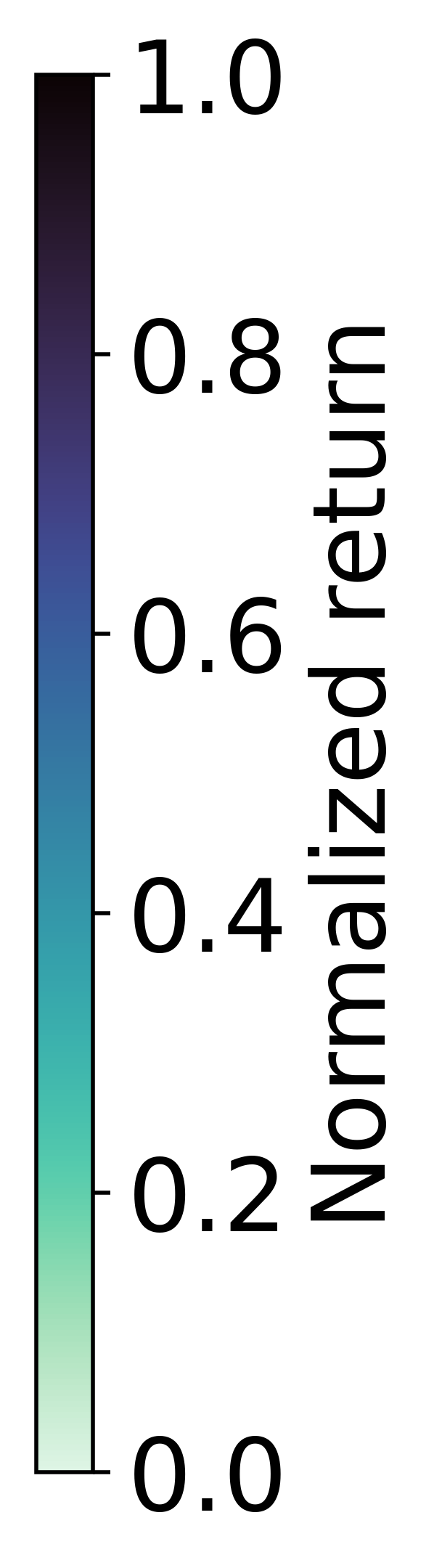}
    % \subfigure[walker]{         \includegraphics[width=0.29\textwidth]{figure/as_tsne/walker2d_medium-replay.png}
    %                             \includegraphics[width=0.29\textwidth]{figure/as_tsne/walker2d_medium.png}
    %                             \includegraphics[width=0.29\textwidth]{figure/as_tsne/walker2d_medium-expert.png}
    %                             \includegraphics[width=0.07\textwidth]{figure/as_tsne/walker2d_heatmap_legend.png}      } \\
    % \subfigure[halfcheetah]{    \includegraphics[width=0.29\textwidth]{figure/as_tsne/halfcheetah_medium-replay.png}
    %                             \includegraphics[width=0.29\textwidth]{figure/as_tsne/halfcheetah_medium.png}
    %                             \includegraphics[width=0.29\textwidth]{figure/as_tsne/halfcheetah_medium-expert.png}
    %                             \includegraphics[width=0.07\textwidth]{figure/as_tsne/halfcheetah_heatmap_legend.png}   }
    % \vspace{-10pt}
	\caption{
	    t-SNE visualization of the embedding space learned with OPPO-a in Hopper environment.
	}
	\label{fig:ablation-q4}
	\vspace{-5pt}
\end{figure*}

\begin{table}[ht]
    \small
    \centering
    % \vspace{-10pt}
    \caption{Ablation study of one-step paradigm in Medium-Replay Dataset}
    \label{tab:ablation-q4}
    \begin{tabular}{lrrrr}
    \toprule
    \multicolumn{1}{c}{{\textbf{Task}}}       & \multicolumn{1}{c}{\textbf{OPPO}}   & \multicolumn{1}{c}{\textbf{OPPO-a}}       \\ \midrule
    Hopper                                    & \textbf{88.9  $\pm$ 2.3}            & 78.3 $\pm$ 7.1                            \\
    Walker                                    & \textbf{71.7  $\pm$ 4.4}            & 66.3 $\pm$ 1.6                            \\
    HalfCheetah                               & \textbf{39.8  $\pm$ 0.2}            & 39.6 $\pm$ 0.1                            \\ \midrule
    \multicolumn{1}{c}{\textbf{Sum}}          & \textbf{200.4}                      & 184.2                                     \\ \bottomrule
    \end{tabular}
    % \vspace{-10pt}
\end{table}

\subsection{Benefits of One-step Offline PbRL}\label{sec:one-step}

We conduct an ablation study to analyze the benefit of iterating $\mathcal{L}_\text{HIM}$ and $\mathcal{L}_\text{PM}$ (for updating $I_\theta$) in an one-step paradigm.
Firstly, we remove $I_\theta$ from $\partial \mathcal{L}_{\text{PM}} / \partial \theta $ and only keep the optimal embedding $\bz^*$ to be updated in Eq.\ref{eq:obj-oppo-3}.
Then, we continue to visualize the embedding $\bz$-space for this ablation setting (OPPO-a), the t-SNE visualization shown in {Fig.\ref{fig:ablation-q4}}.
By comparing Fig.\ref{fig:ablation-q4} to Fig.\ref{fig:tsne-q1}, we can see that the preference relationship in the embedding space (learned with OPPO-a) is all shuffled.
In a less expressive $\bz$-space, it is challenging to model the preference and find the optimal $\bz^*$.
Further, as shown in Table~\ref{tab:ablation-q4}, the comparison results of medium-replay tasks demonstrate that such an ablation does cause the performance degradation.

%Fundamentally, $\mathcal{L}_{\text{PM}}$ helps the encoder to learn a more informative and visually interpretable $\bz$-space. Removing the preference loss leads to a $\bz$-space in which the embeddings are evenly distributed in a mixing style.However, it is also worth noting that the performance of these modified PDT, i.e. their benchmarked peroformance in D4RL, is not affected much by this uninformative z-space, as shown in \Cref{table:pdt_with_pdf_modified}. We attribute this to the effectiveness of searching phase, in which our method is still able to find a meaningful $\bz^*$ in a less expressive z-space. This is also justified from t-SNE(\Cref{fig:CR}) as the red cross(position of $\bz^*$) locates just in the point of deep color.

\begin{table}[ht]
    % \arrayrulecolor{red}
    \centering
    % \vspace{-10pt}
    \caption{Ablation study of feedback quantity}\label{tab:ablation-feedback}
    \resizebox{0.48\textwidth}{!}{
    \begin{tabular}{lrrrr}
    \toprule
    \multicolumn{1}{c}{\textbf{Dataset}} & \multicolumn{1}{c}{50k} & \multicolumn{1}{c}{1k}      & \multicolumn{1}{c}{500} \\ \midrule
    Medium-Expert                        & 108.0  $\pm$ 5.1        & 102.9    $\pm$     3.2      & 104.9  $\pm$  4.1       \\
    Medium                               & 86.3   $\pm$ 3.2        & 90.8     $\pm$     2.0      & 77.5   $\pm$  12.8      \\
    Medium-Replay                        & 88.9   $\pm$ 2.3        & 60.4     $\pm$     3.0      & 68.5   $\pm$  22.8      \\ \midrule
    Sum                                  & 283.1                   & 254.2                       & 250.9                   \\ \bottomrule
    \end{tabular}}
    % \vspace{-10pt}
\end{table}

\subsection{Performance with Different Amount of Preference Feedback}\label{sec:diff-amount feedback}
For the Hopper task, we evaluate the impact of different amounts of preference labels on the performance of OPPO and show the results in Table \ref{tab:ablation-feedback}.
Specifically, OPPO is evaluated using the labels amount from {50k, 1k, 500}, on the dataset from {Medium-Expert, Medium, Medium Replay}.
As illustrated in the Table \ref{tab:ablation-feedback}, OPPO performs the best when given 50k preference labels and achieves a total normalized score of 283.1 among the three datasets.
% When the feedback amount decreases to 3k, the performance decreases correspondingly.
However, the performance decreases at around 250 for feedback amount decreases to 1k and 500.
Therefore, OPPO is robust to the variation in terms of the amount of preference feedback used for training.

% \begin{table}[t]
%     % \arrayrulecolor{red}
%     \centering
%     \begin{tabular}{lrrrr}
%     \toprule
%     \multicolumn{1}{c}{\textbf{Dataset}} & \multicolumn{1}{c}{\textbf{50k}} & \multicolumn{1}{c}{3k} & \multicolumn{1}{c}{\textbf{1k}} & \multicolumn{1}{c}{500} \\ \midrule
%     Medium-Expert                        & 108.0  $\pm$ 5.1                 & 92.1   $\pm$   9.2     & 102.9    $\pm$     3.2          & 104.9  $\pm$  4.1       \\
%     Medium                               & 86.3   $\pm$ 3.2                 & 73.5   $\pm$   14.8    & 90.8     $\pm$     2.0          & 77.5   $\pm$  12.8      \\
%     Medium-Replay                        & 88.9   $\pm$ 2.3                 & 66.2   $\pm$   23.3    & 60.4     $\pm$     3.0          & 68.5   $\pm$  22.8      \\ \midrule
%     Sum                                  & 283.1                            & 231.8                  & 254.2                           & 250.9                   \\ \bottomrule
%     \end{tabular}
%     \caption{Ablation study}\label{tab:ablation-feedback}
% \end{table}

\begin{table}[ht]
    \centering
    \caption{
        Performance OPPO with preference from real human, where results are reported over 3 seeds.
        }
    \label{tab:humanteacher}
    \resizebox{0.48\textwidth}{!}{
    \begin{tabular}{llrrr}
    \toprule
    \multicolumn{1}{c}{\textbf{Environment}}    & \multicolumn{1}{c}{\textbf{Dataset}} & \multicolumn{1}{c}{\textbf{IQL+$r$}} & \multicolumn{1}{c}{\textbf{IQL+PT}} & \multicolumn{1}{c}{\textbf{OPPO}}   \\ \midrule
    Hopper                                      & Medium-Expert                        & 73.6  $\pm$ 41.5                     & 69.0  $\pm$ 33.9                    & \textbf{107.8 $\pm$ 1.6}            \\
                                                & Medium-Replay                        & 83.1  $\pm$ 15.8                     & 84.5  $\pm$ 4.1                     & \textbf{93.1  $\pm$ 1.3}            \\ \midrule
    Walker                                      & Medium-Expert                        & 107.8 $\pm$ 2.0                      & \textbf{110.1 $\pm$ 0.2}            & 106.4 $\pm$ 1.1                     \\
                                                & Medium-Replay                        & 73.1  $\pm$ 8.1                      & 71.3  $\pm$ 10.3                    & \textbf{74.9  $\pm$ 0.7}            \\ \midrule
    \multicolumn{2}{c}{\textbf{Locomotion Sum}}                                        & 337.47                               & 334.9                               & \textbf{382.2}                      \\ \midrule
    Lift                                        & Proficient-Human                     & \textbf{96.8 $\pm$ 1.8}              & 91.8 $\pm$ 5.9                      & 94.7 $\pm$ 1.2                      \\
                                                & Multi-Human                          & 86.8 $\pm$ 2.8                       & 86.8 $\pm$ 6.0                      & \textbf{98.7 $\pm$ 2.3}             \\ \midrule
    Can                                         & Proficient-Human                     & 74.5 $\pm$ 6.8                       & 69.7 $\pm$ 5.9                      & \textbf{75.3 $\pm$ 10.1}            \\
                                                & Multi-Human                          & 56.3 $\pm$ 8.8                       & 50.5 $\pm$ 6.5                      & \textbf{86.7 $\pm$ 12.7}            \\ \midrule
    \multicolumn{2}{c}{\textbf{Robosuite Sum}}                                         & 314.3                                & 298.7                               & \textbf{355.3}                      \\ \bottomrule
    \end{tabular}}
\end{table}

\subsection{Performance of OPPO on Benchmark Tasks with Real Human Teacher} \label{sec:real_human_teacher}
We have conducted additional experiments using real human-labeled data on the Hopper and Walker tasks.
The human preferences we used are obtained from the open-source dataset of PT~\citep{kim2023preference}, which is collected from actual human familiar with robotic tasks.
Also, we have carried out experiments on the Robomimic dataset~\citep{mandlekar2022matters}, which offers a set of offline datasets on 7-DoF robot manipulation domains.
In our experiments, we test our method on two tasks (Lift and Can), where the offline data are collected by either one proficient human teleoperators (Proficient-Human) or multiple human teleoperators with varying proficiency (Multi-Human), and the preference labels are also labeled by real human.
In Table \ref{tab:humanteacher}, we compare OPPO to IQL+$r$~\citep{kostrikov2022offline} and IQL+PT~\citep{kim2023preference}.
We find that our method outperforms IQL+PT in most tasks, and even achieves competitive or better preformance than IQL+$r$.

To sum up, through six experiments and the visualization of the results, we demonstrate that the $\bz$-space learned by the encoder is informative and visually interpretable.
Besides, the ablation study proves that a preference-guided embedding space of context could improve task performance asymptotically by a non-neglectable margin.
Moreover, OPPO can find an embedding to represent the context of the optimal trajectory, where the resulting trajectory is better than any offline trajectory in the dataset.
Last but not least, in the offline setting with environment interaction disabled, our paradigm can acquire the optimal behaviors using binary preference labels between sub-optimal trajectories.
As shown in the experiment results, OPPO achieves a competitive performance over DT trained using either true rewards or pseudo rewards.

\section{Conclusion}

This paper introduces offline preference-guided policy optimization (OPPO), a one-step offline PbRL paradigm.
Unlike the previous PbRL approaches that learn policy from a pseudo-reward function (learning a separate reward function is a prerequisite), OPPO directly optimizes the policy in a high-level embedding space.
To enable that, we propose an offline hindsight information matching (HIM) objective and a preference modeling objective.
Empirically, we show that iterating the above two objectives can produce meaningful and preference-aligned embeddings of context.
Moreover, conditioned on the learned optimal context, our HIM-based contextual policy can achieve competitive performance on standard offline (PbRL) tasks.

% \section*{Software and Data}

% If a paper is accepted, we strongly encourage the publication of software and data with the
% camera-ready version of the paper whenever appropriate. This can be
% done by including a URL in the camera-ready copy. However, \textbf{do not}
% include URLs that reveal your institution or identity in your
% submission for review. Instead, provide an anonymous URL or upload
% the material as ``Supplementary Material'' into the CMT reviewing
% system. Note that reviewers are not required to look at this material
% when writing their review.

\section*{Acknowledgements}
This work was supported by STI 2030—Major Projects (2022ZD0208800), and NSFC General Program (Grant No. 62176215).

% \textbf{Do not} include acknowledgements in the initial version of
% the paper submitted for blind review.

% If a paper is accepted, the final camera-ready version can (and
% probably should) include acknowledgements. In this case, please
% place such acknowledgements in an unnumbered section at the
% end of the paper. Typically, this will include thanks to reviewers
% who gave useful comments, to colleagues who contributed to the ideas,
% and to funding agencies and corporate sponsors that provided financial


\bibliography{OPPO}
\bibliographystyle{icml2023}

%%%%%%%%%%%%%%%%%%%%%%%%%%%%%%%%%%%%%%%%%%%%%%%%%%%%%%%%%%%%%%%%%%%%%%%%%%%%%%%
%%%%%%%%%%%%%%%%%%%%%%%%%%%%%%%%%%%%%%%%%%%%%%%%%%%%%%%%%%%%%%%%%%%%%%%%%%%%%%%
% APPENDIX
%%%%%%%%%%%%%%%%%%%%%%%%%%%%%%%%%%%%%%%%%%%%%%%%%%%%%%%%%%%%%%%%%%%%%%%%%%%%%%%
%%%%%%%%%%%%%%%%%%%%%%%%%%%%%%%%%%%%%%%%%%%%%%%%%%%%%%%%%%%%%%%%%%%%%%%%%%%%%%%
\newpage
\appendix
\onecolumn

% You can have as much text here as you want. The main body must be at most $8$ pages long.
% For the final version, one more page can be added.
% If you want, you can use an appendix like this one, even using the one-column format.

\section{Appendix}

\subsection{Implementation details}

\subsubsection{Codebase.}
Our code is based on Decision Transformer\footnote{\url{https://github.com/kzl/decision-transformer}}, and our implementation of OPPO is available at:

\centerline{\textcolor{magenta}{\textbf{\url{https://github.com/bkkgbkjb/OPPO}}}}

\subsubsection{OpenAI Gym.}
% Our code is based on the Huggingface Transformers library~\citep{wolf-etal-2020-transformers}.
We choose the OpenAI Gym continuous control tasks from the D4RL benchmark~\citep{fu2020d4rl}.
The different dataset settings are described below.
\begin{itemize}
	\item Medium: 1 million timesteps generated by a "medium" policy that achieves approximately one-third of the score of an expert policy.
	\item Medium-Replay: the replay buffer of an agent trained to the performance of a medium policy (approximately 25k-400k timesteps in our environments).
	\item Medium-Expert: 1 million timesteps generated by the medium policy concatenated with 1 million timesteps generated by an expert policy.
\end{itemize}
For details of these environments and datasets, please refer to D4RL for more information.

\subsubsection{Hyperparameters}\label{hyperparameters}

During the offline HIM phase, we weighted sum all three losses as in Eq.\ref{eq:total-loss} (with ratios listed in \Cref{hyper-loss}) and perform backpropagation, while in Preference Modeling phase, only $\mathcal{L}_{\text{PM}}$ is computed and backpropagated.

\begin{table}[ht]
    \centering
    \caption{Hyperparameters of coefficients of combined losses during Offline HIM.}
    \label{hyper-loss}
    \begin{tabular}{ll}
    \toprule
    \multicolumn{1}{c}{\textbf{Hyperparameter}} & \multicolumn{1}{c}{\textbf{Value}}    \\ \midrule
    $\alpha$                                    & 0.25 for halfcheetah-medium-expert    \\
                                                & 0.5 for others                        \\
    $\beta$                                     & 0.05 for halfcheetah-medium-expert    \\
                                                & 0.1 for others                        \\ \bottomrule
    \end{tabular}
\end{table}

Our hyperparameters on all tasks are shown below in \Cref{hyper-z} and \Cref{hyper}.
Models were trained for $10^{5}$ gradient steps using the AdamW optimizer~\citet{loshchilov2017decoupled} following PyTorch defaults.
% We search over $K\in{5, 20, 100}$ (which all yielded similar results), and did not search over $\hat{R}$ (we conditioned on expert performance for all tasks, except in Cheetah where the dataset did not contain expert data).

\begin{table}[ht]
    \centering
    \caption{Hyperparameters of $\bz^*$ searching for OpenAI Gym experiments.}
    \label{hyper-z}
    \begin{tabular}{ll}
    \toprule
    \multicolumn{1}{c}{\textbf{Hyperparameter}} & \multicolumn{1}{c}{\textbf{Value}}       \\ \midrule
    Number of dimensions                        & 8 for halfcheetah                        \\
                                                & 16 for others                            \\
    Amount of feedback                          & 50k                                      \\
    Type of optimizer                           & AdamW                                    \\
    Learning rate                               & $10^{-2}$ for halfcheetah-medium-expert  \\
                                                & $10^{-3}$ for others                     \\
    Weight decay                                & $10^{-4}$                                \\
    Margin                                      & 1                                        \\ \bottomrule
    \end{tabular}
\end{table}

\begin{table}[ht]
    \centering
    \caption{Hyperparameters of Transformer for OpenAI Gym experiments.}
    \label{hyper}
    \begin{tabular}{ll}
    \toprule
    \multicolumn{1}{c}{\textbf{Hyperparameter}} & \multicolumn{1}{c}{\textbf{Value}}              \\ \midrule
    Number of layers                            & 3                                               \\
    Number of attention heads                   & 2 for encoder transformer                       \\
                                                & 1 for decision transformer                      \\
    Embedding dimension                         & 128                                             \\
    Nonlinearity function                       & ReLU                                            \\
    Batch size                                  & 64                                              \\
    context length K                            & 20                                              \\
    Dropout                                     & 0.1                                             \\
    Learning rate                               & $10^{-4}$                                       \\
    Grad norm clip                              & 0.25                                            \\
    Weight decay                                & $10^{-4}$                                       \\
    Learning rate decay                         & Linear warmup for first $10^{5}$ training steps \\ \bottomrule
    \end{tabular}
\end{table}

\subsubsection{Computational resources.}
The experiments were run on a computational cluster with 20x GeForce RTX 2080 Ti, and 4x NVIDIA Tesla V100 32GB for about 20 days.

\subsection{Additional results}

\subsubsection{More visualization results on $\bz$-space.}

\begin{figure}[ht]
	\centering
    % \vspace{-10pt}
    % 	% \includegraphics[width=0.6\textwidth]{exp_fig/baseline/legend.pdf} \\
    	% \vspace{-5pt}
        % \subfigure[hopper]{         \includegraphics[width=0.29\textwidth]{figure/tsne/hopper_medium-replay.png}
        %                             \includegraphics[width=0.29\textwidth]{figure/tsne/hopper_medium.png}
        %                             \includegraphics[width=0.29\textwidth]{figure/tsne/hopper_medium-expert.png}
        %                             \includegraphics[width=0.07\textwidth]{figure/tsne/hopper_heatmap_legend.png}       }
    \subfigure[walker]{         \includegraphics[width=0.28\textwidth]{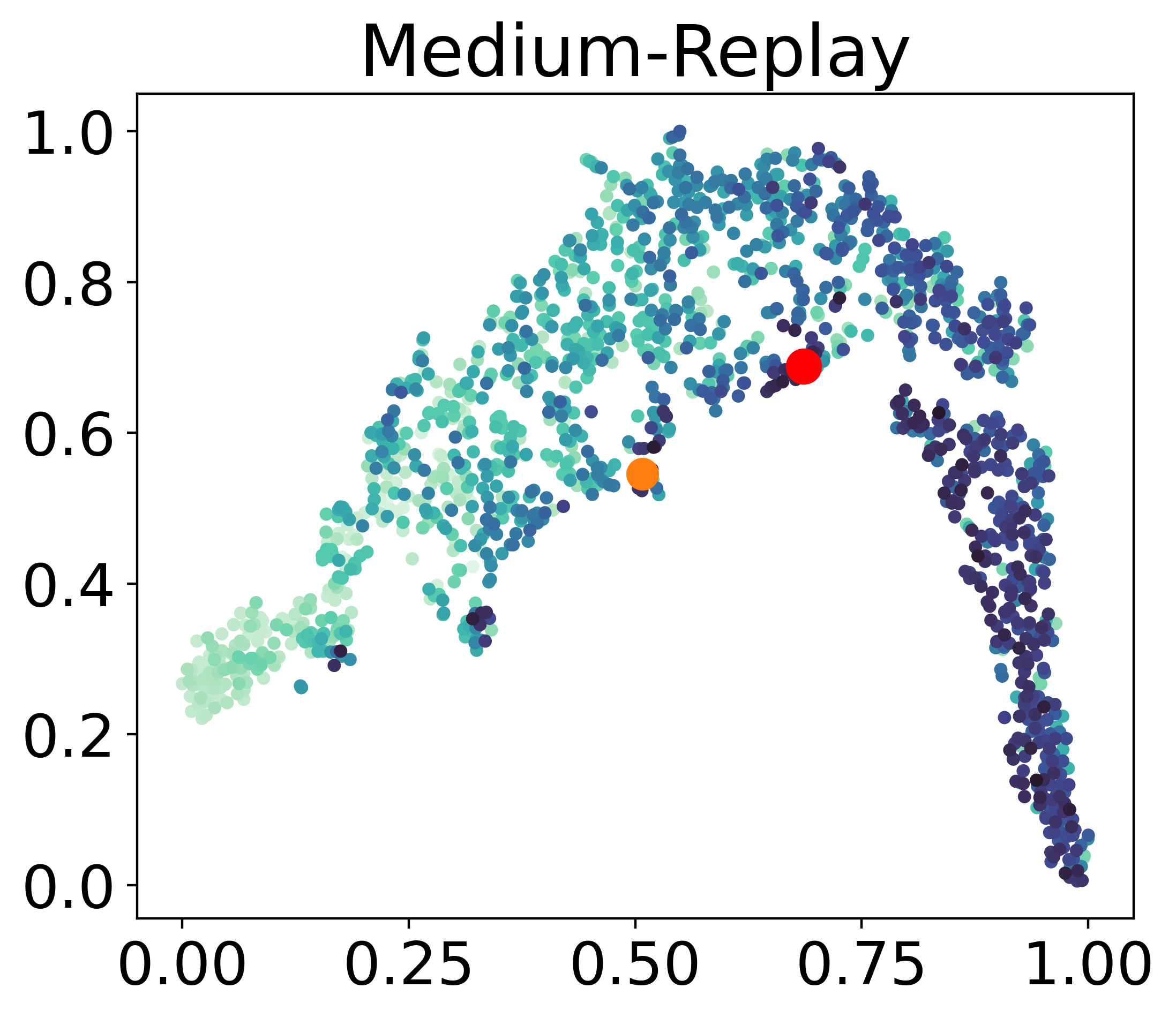}
                                \includegraphics[width=0.28\textwidth]{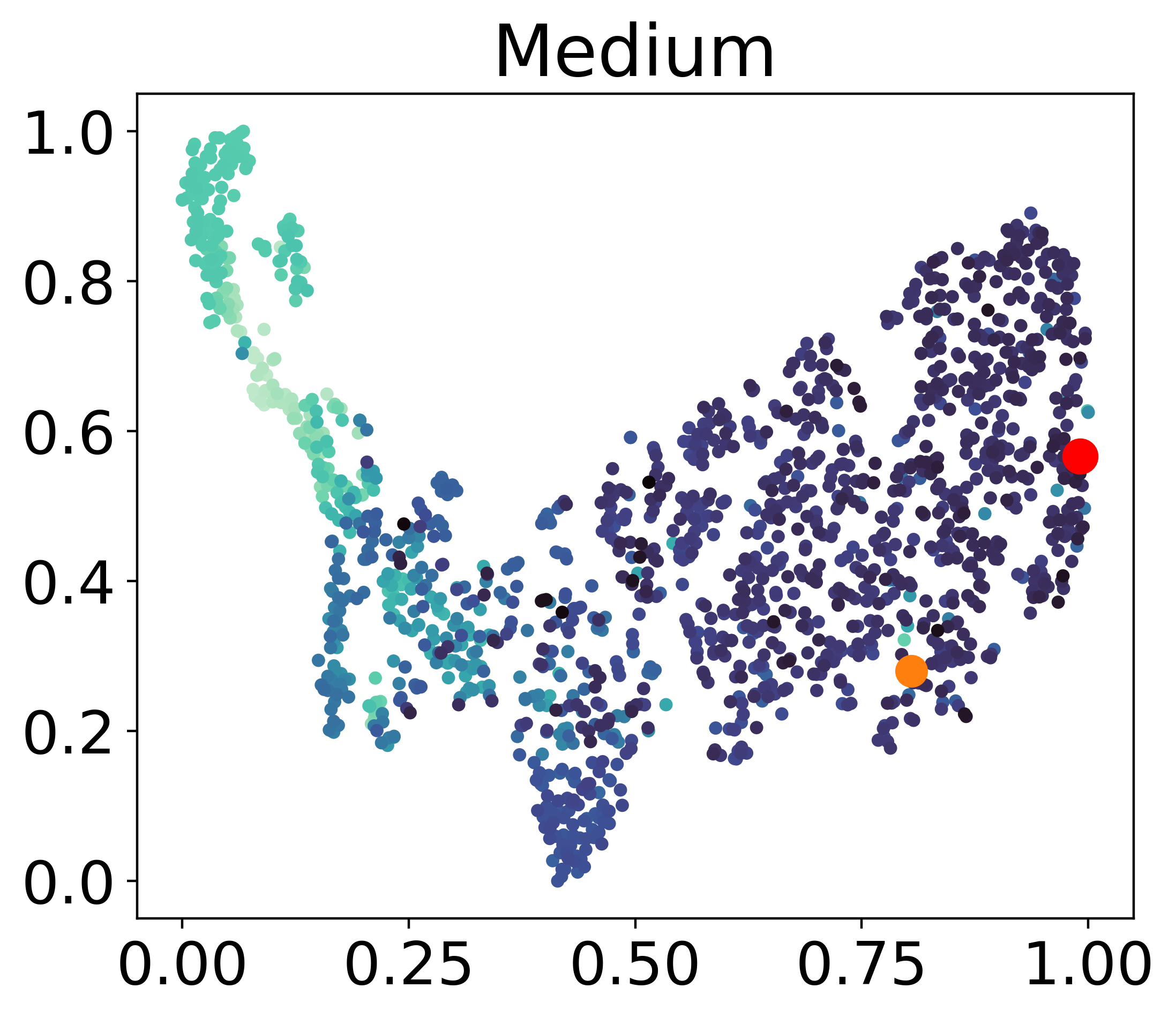}
                                \includegraphics[width=0.28\textwidth]{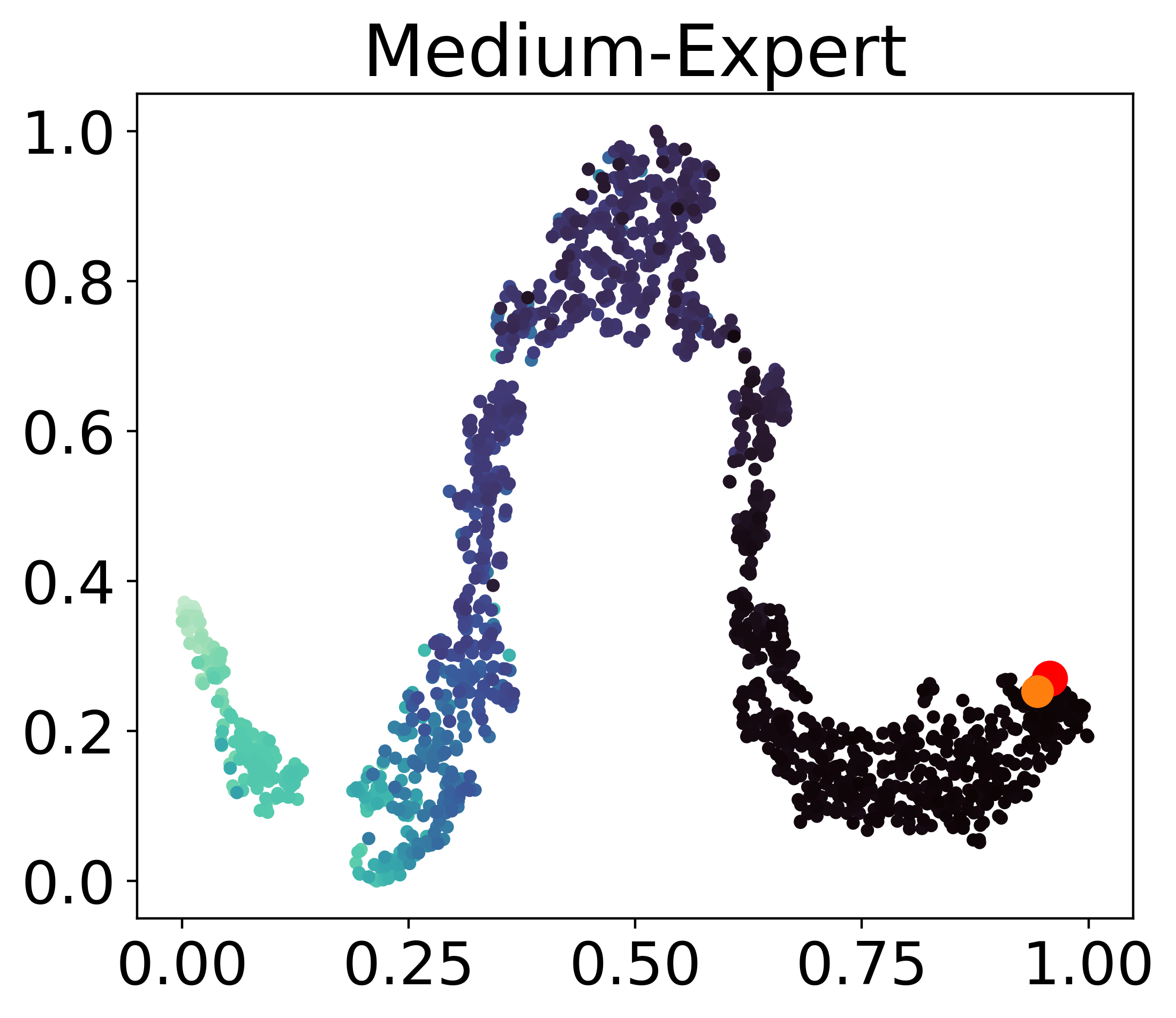}
                                \includegraphics[width=0.068\textwidth]{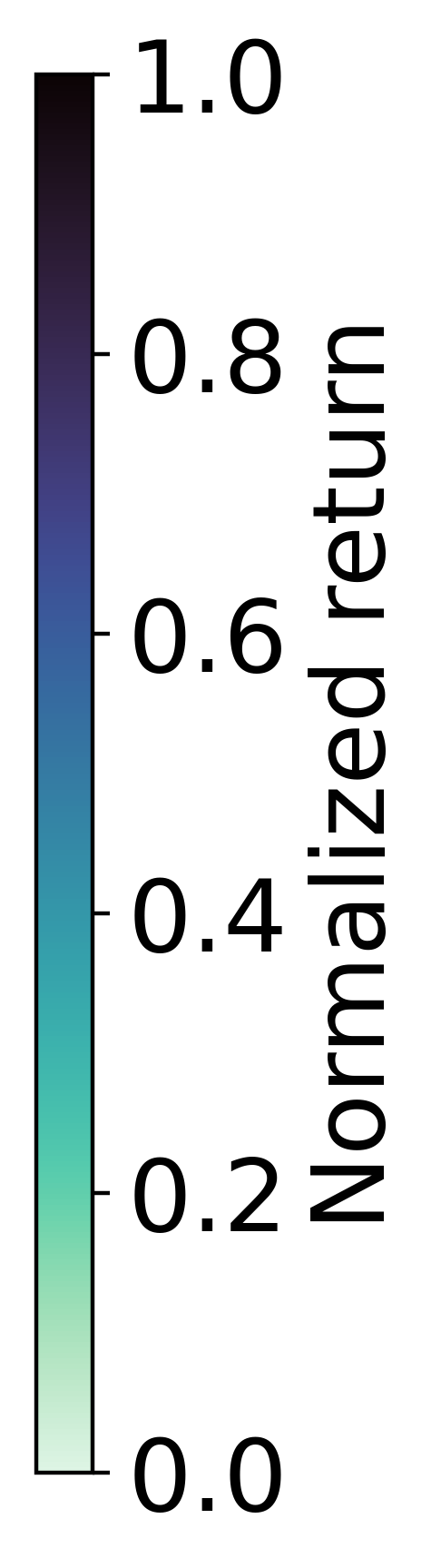}     } \\
    \subfigure[halfcheetah]{    \includegraphics[width=0.28\textwidth]{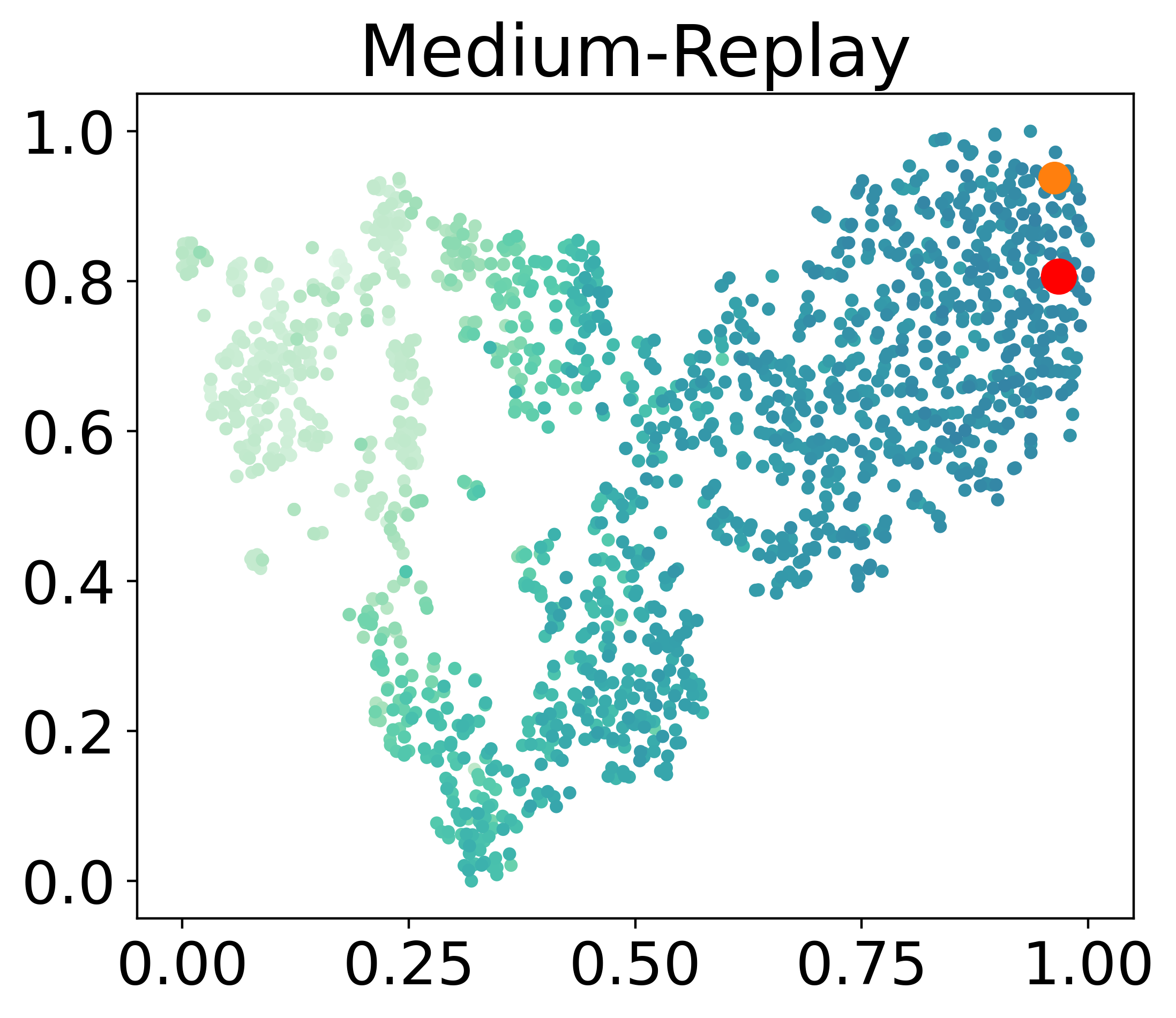}
                                \includegraphics[width=0.28\textwidth]{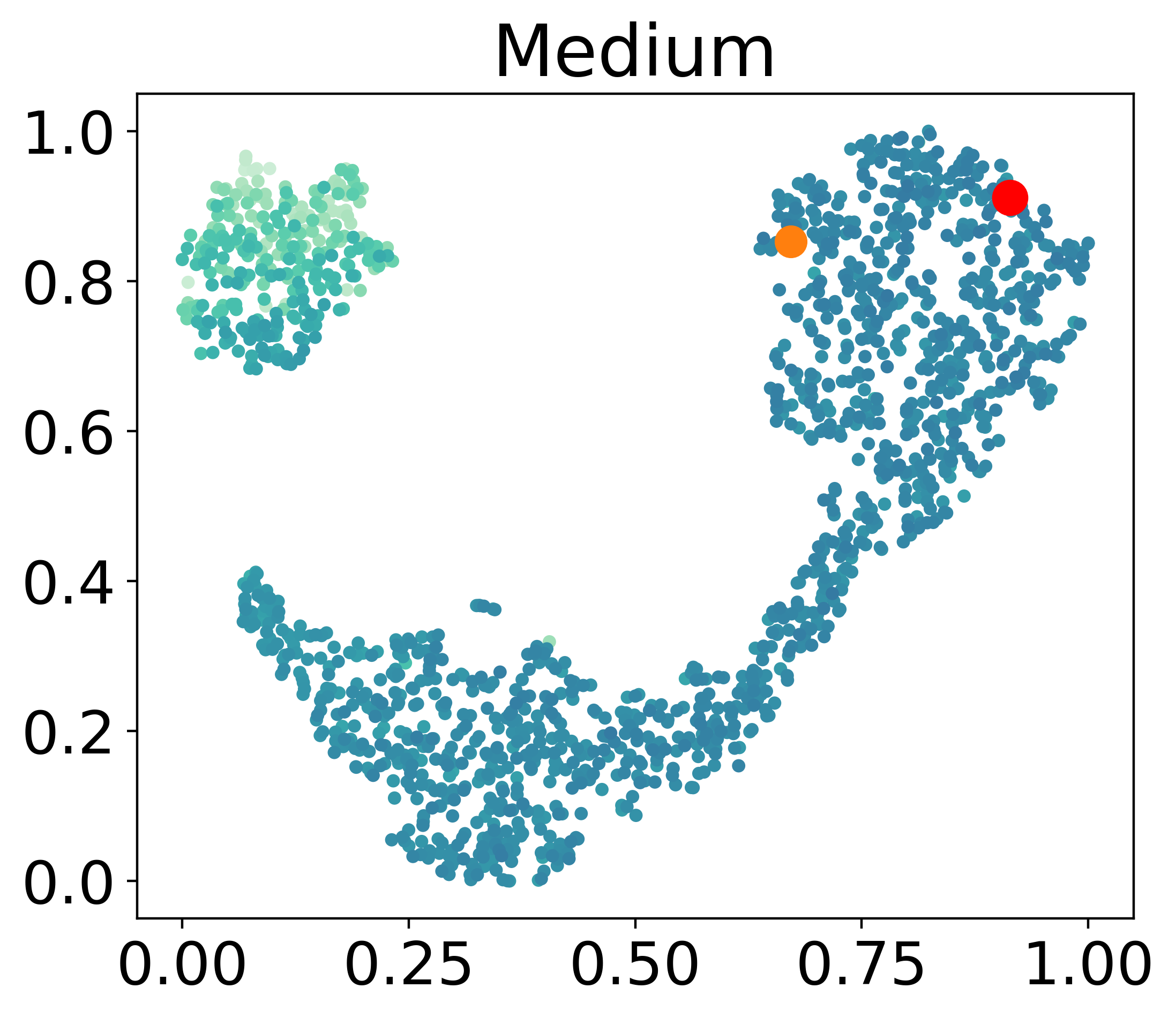}
                                \includegraphics[width=0.28\textwidth]{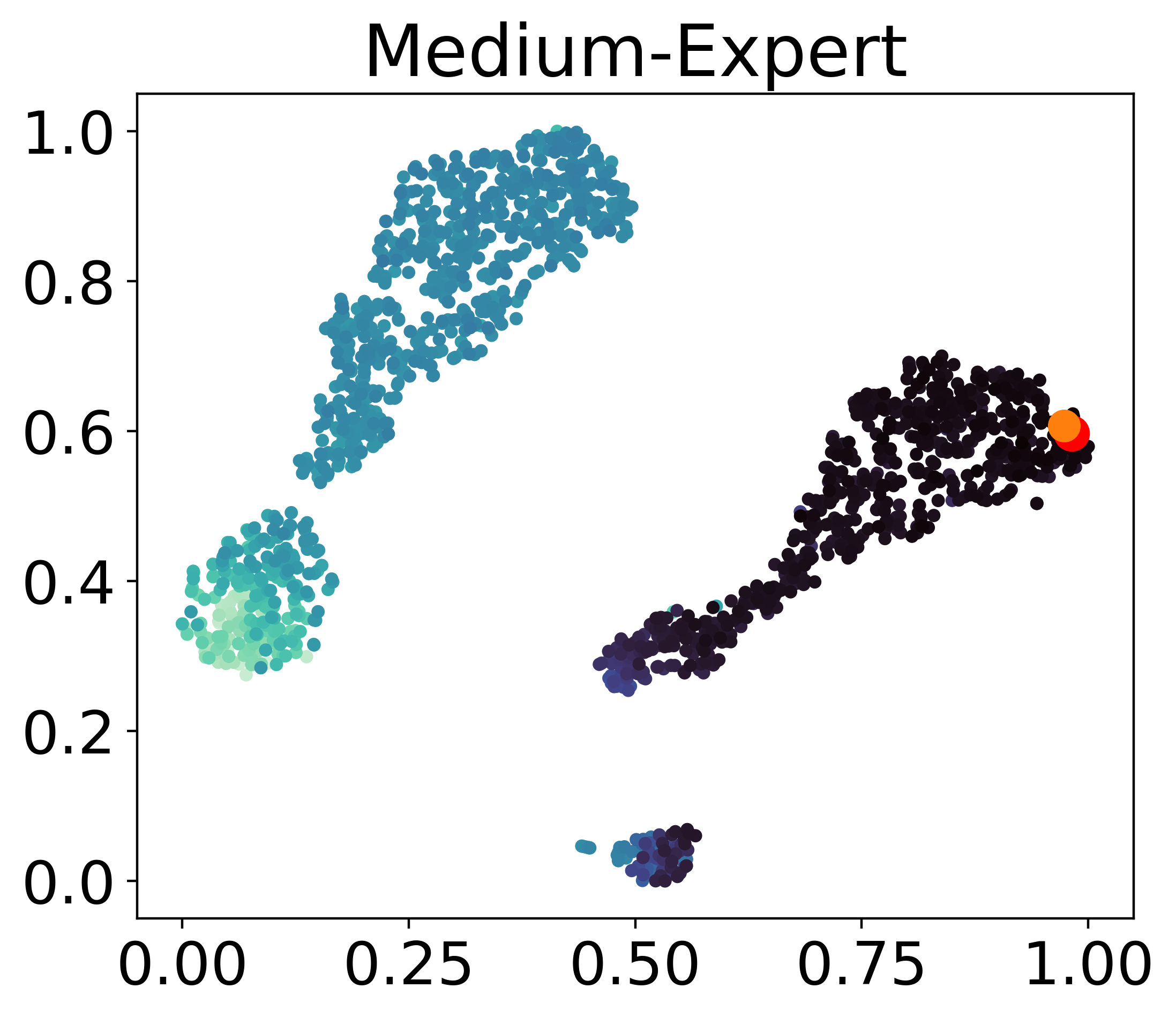}
                                \includegraphics[width=0.068\textwidth]{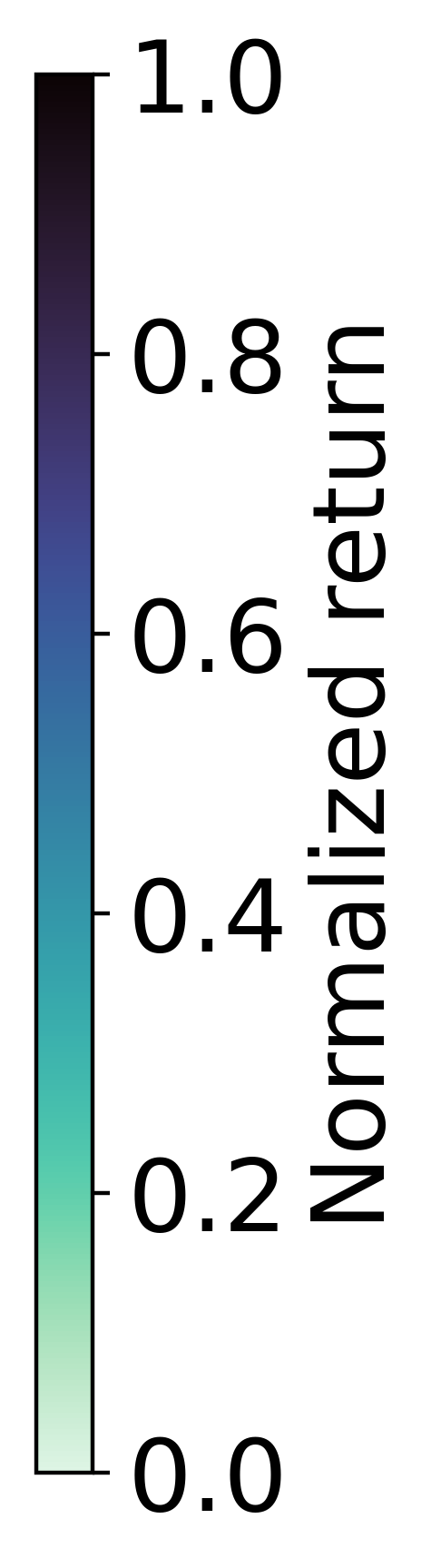}}
	\caption{
        t-SNE of OPPO in Qalker and HalfCheetah including the embedding of random trajectories in $\mathcal{D}$, the colors of the points represent the normalized return of the corresponding trajectory $\tau$.
	}
	\label{fig:tsne1-appendix}
	\vspace{-10pt}
\end{figure}

We further show the t-sne results of OPPO in \ref{fig:tsne1-appendix} with the setting described in Section \ref{sec:zspace} in Walker and HalfCheetah environments.

% To answer this question, we design an ablation study that removes preference loss from the optimization objective of the Encoder Transformer (Note: we still enforce preference loss in the search of $\bz^*$).
% During the evaluation phase, we once again utilize T-SNE to visiualize the $\bz$-space learned without $\mathcal{L}_{\text{PM}}$, and obtain \Cref{fig:CR}.

% By training this modified PDT on 9 environments, then use well-trained Encoder-Transformer of it to obtain a  analysis similar to section 1, we could visualize into the

Our primary purpose of using t-SNE is for visualization, to illustrate the structure of the learned $\bz$-space in a more intuitive rather than quantitative manner.
While t-SNE results are known to be hyper-parameter dependent~\citep{wattenberg2016how}, we have included in Table \ref{tab:Euclidean} that listing the euclidean distances of $\bz^*$ from $\bz^{**}$ on different tasks.

\begin{table}[ht]
    \small
    \centering
    \caption{Euclidean distances of $\bz^*$ from $\bz^{**}$ on different tasks.}
    \label{tab:Euclidean}
    \begin{tabular}{llrcr}
    \toprule
    \multicolumn{1}{c}{\textbf{Environment}} & \multicolumn{1}{c}{\textbf{Dataset}} & \multicolumn{1}{c}{\textbf{$||\mathbf{z}^*-\mathbf{z}^{**}||_2$}} & \makecell{\textbf{{[}Min, Lower quartile, Median,} \\ \textbf{Upper quartile, Max{]}}}  & \makecell{\textbf{Percentile} \\ \textbf{Rank (PR)}} \\ \midrule
    Hopper                                   & Medium-Expert                        & 12.68                                                             & {[}12.40, 13.29, 14.19, 15.01, 16.37{]}                                                 & 97.9\%                                               \\
                                             & Medium                               & 15.50                                                             & {[}13.20, 13.92, 14.81, 15.70, 17.23{]}                                                 & 30.1\%                                               \\
                                             & Medium-Replay                        & 13.33                                                             & {[}12.13, 13.24, 13.98, 14.83, 16.45{]}                                                 & 72.0\%                                               \\ \midrule
    Walker                                   & Medium-Expert                        & 13.16                                                             & {[}12.62, 13.02, 14.13, 15.02, 16.40{]}                                                 & 71.2\%                                               \\
                                             & Medium                               & 12.86                                                             & {[}12.01, 12.84, 13.21, 14.03, 15.79{]}                                                 & 74.1\%                                               \\
                                             & Medium-Replay                        & 14.26                                                             & {[}10.90, 12.39, 13.14, 13.58, 15.18{]}                                                 & 6.7\%                                                \\ \midrule
    HalfCheetah                              & Medium-Expert                        & 10.42                                                             & {[}10.34, 10.82, 12.35, 12.71, 13.85{]}                                                 & 99.7\%                                               \\
                                             & Medium                               & 3.22                                                              & {[}3.19,  3.40,  4.15,  4.54,  5.80{]}                                                  & 99.9\%                                               \\
                                             & Medium-Replay                        & 1.89                                                              & {[}1.53,  1.98,  2.92,  3.69,  4.24{]}                                                  & 79.9\%                                               \\ \bottomrule
    \end{tabular}
\end{table}

To provide some context, we also calculated the distances between the embeddings of trajectories in each dataset and $\bz^*$, then gathered the minimum, lower quartile, median, upper quartile, and maximum values in Table \ref{tab:Euclidean}.
Additionally, we included percentile rankings for the distances between $\bz^*$ and $\bz^{**}$ within each dataset.

The results confirm the intuitions from the t-SNE plots, and the percentile rank maybe more informative.

\subsubsection{More results of ablation study of one-step paradigm}

\begin{figure}[ht]
	\centering
% 	% \includegraphics[width=0.6\textwidth]{exp_fig/baseline/legend.pdf} \\
% 	% \vspace{-5pt}
    % \subfigure[hopper]{         \includegraphics[width=0.29\textwidth]{figure/as_tsne/hopper_medium-replay.png}
    %                             \includegraphics[width=0.29\textwidth]{figure/as_tsne/hopper_medium.png}
    %                             \includegraphics[width=0.29\textwidth]{figure/as_tsne/hopper_medium-expert.png}
    %                             \includegraphics[width=0.07\textwidth]{figure/as_tsne/hopper_heatmap_legend.png}        }
    \subfigure[walker]{         \includegraphics[width=0.29\textwidth]{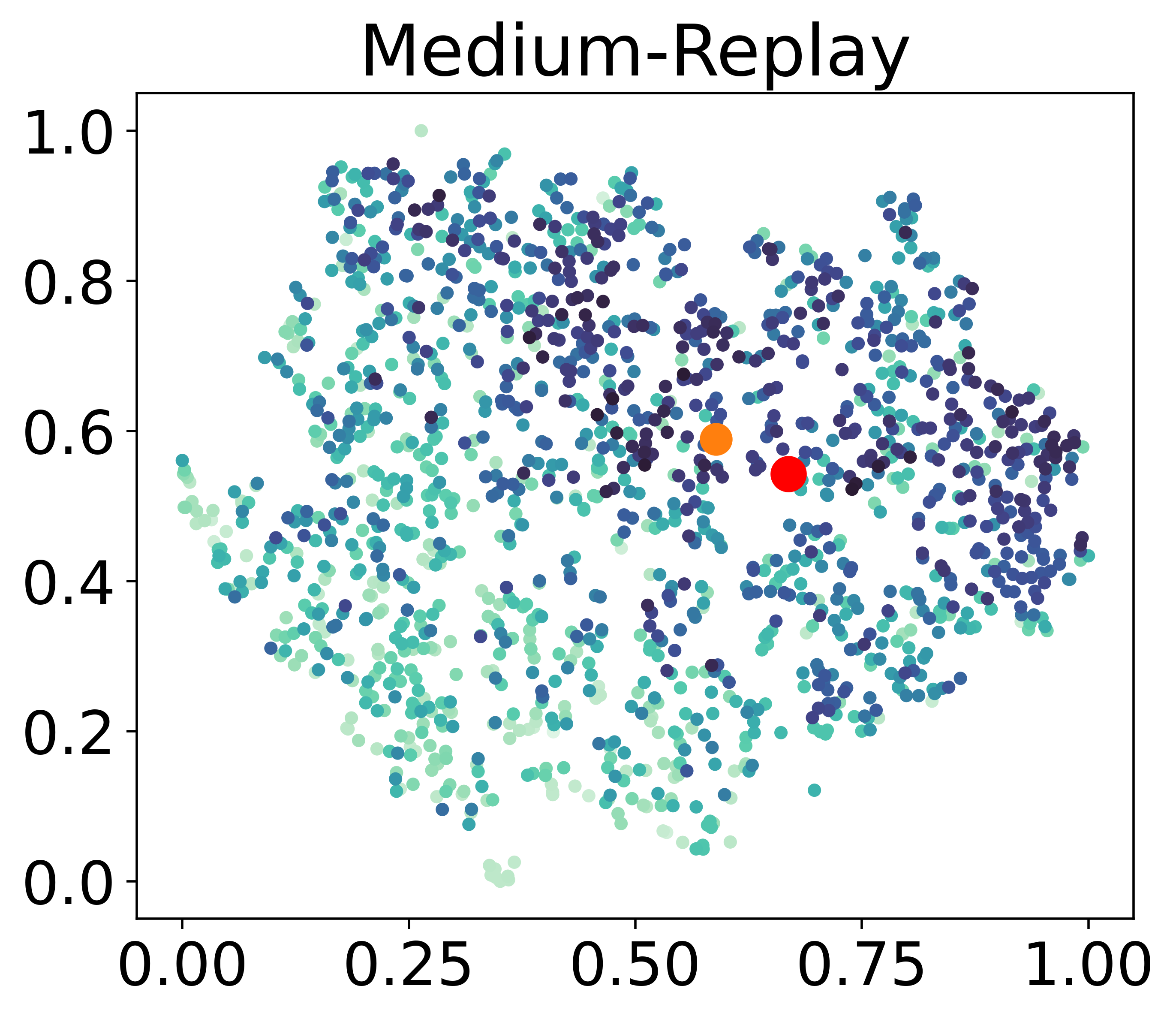}
                                \includegraphics[width=0.29\textwidth]{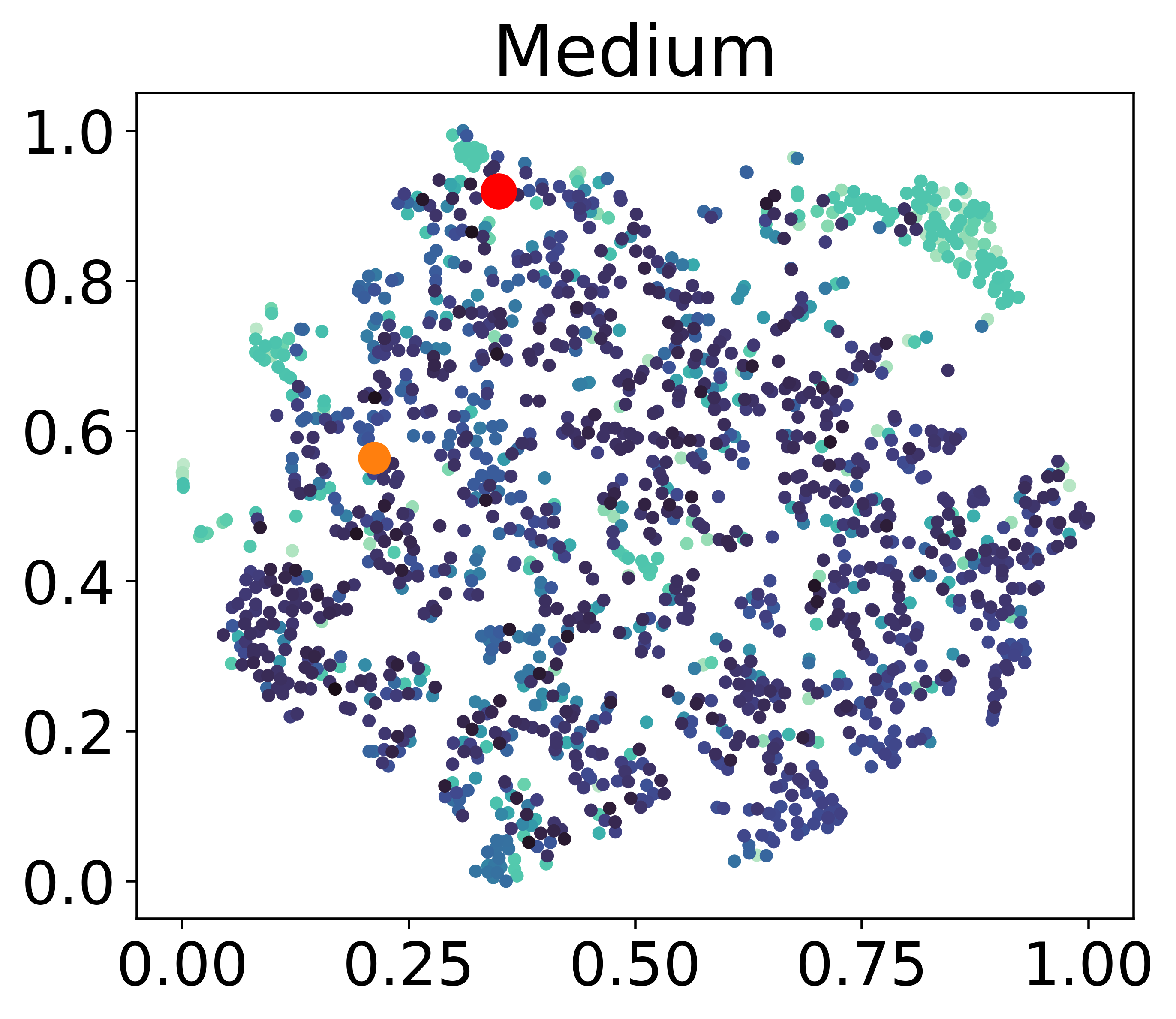}
                                \includegraphics[width=0.29\textwidth]{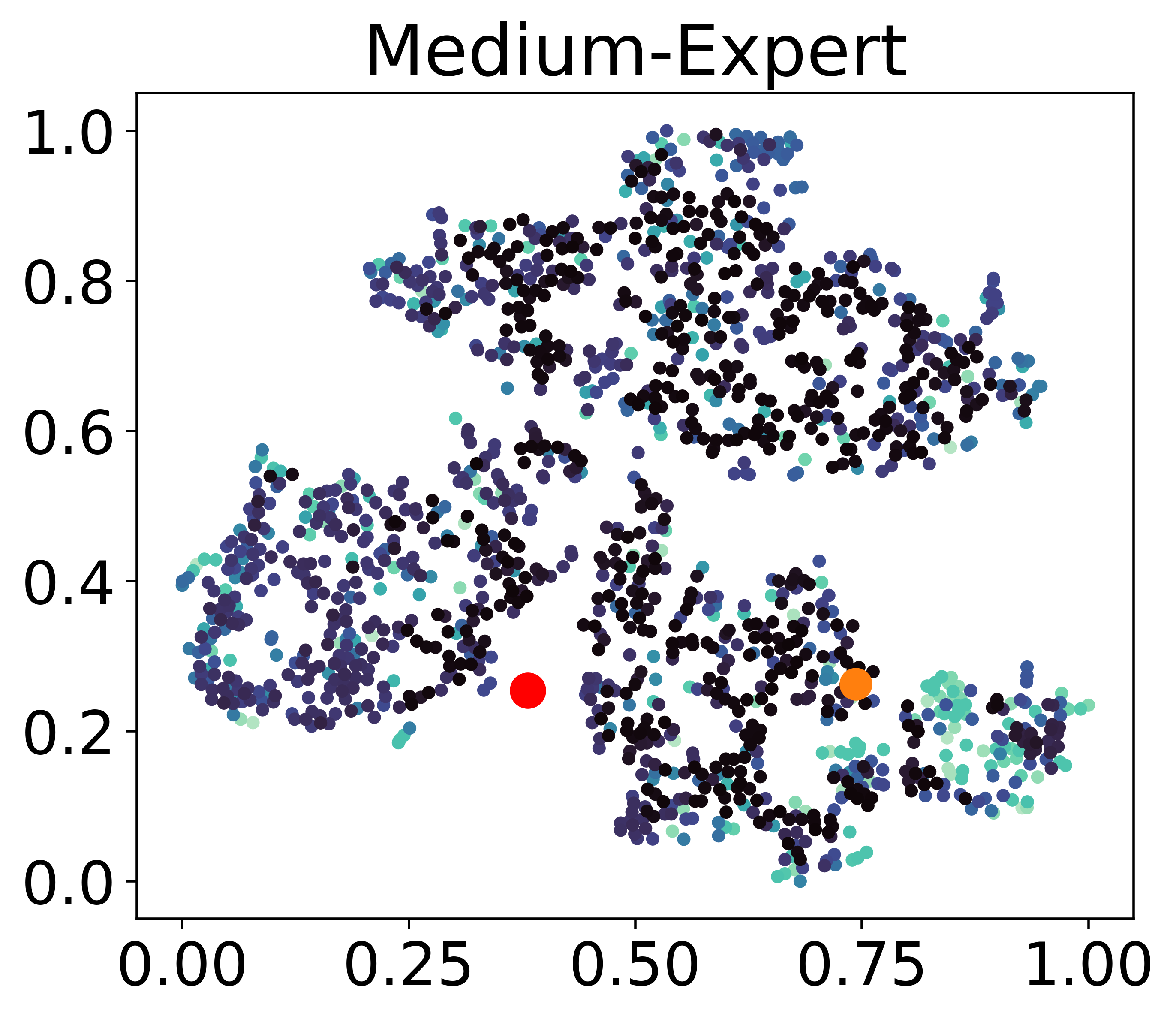}
                                \includegraphics[width=0.07\textwidth]{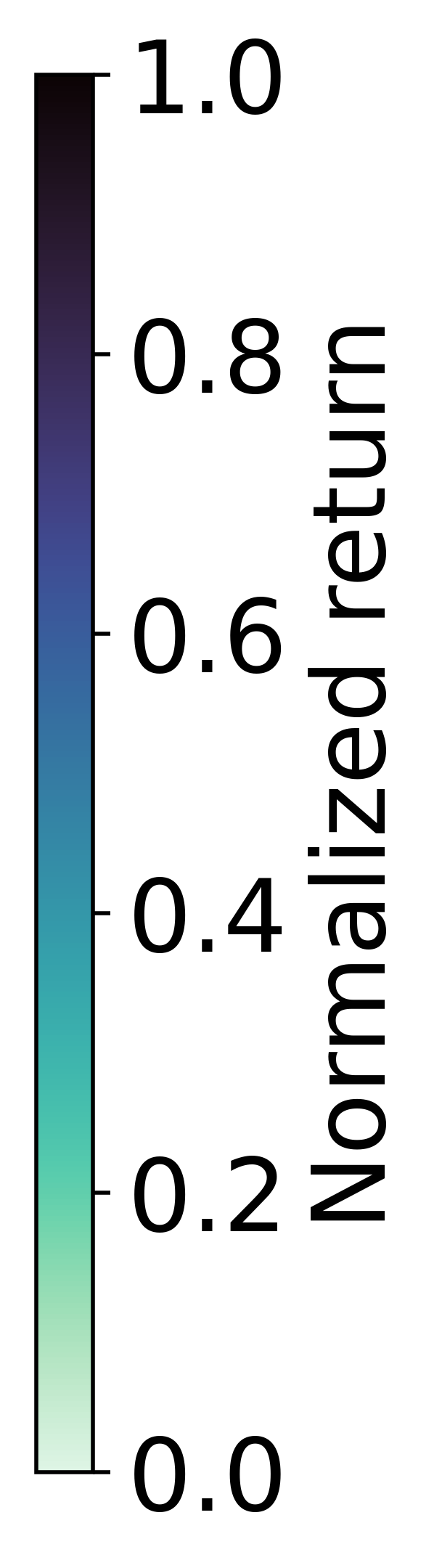}      } \\
    \subfigure[halfcheetah]{    \includegraphics[width=0.29\textwidth]{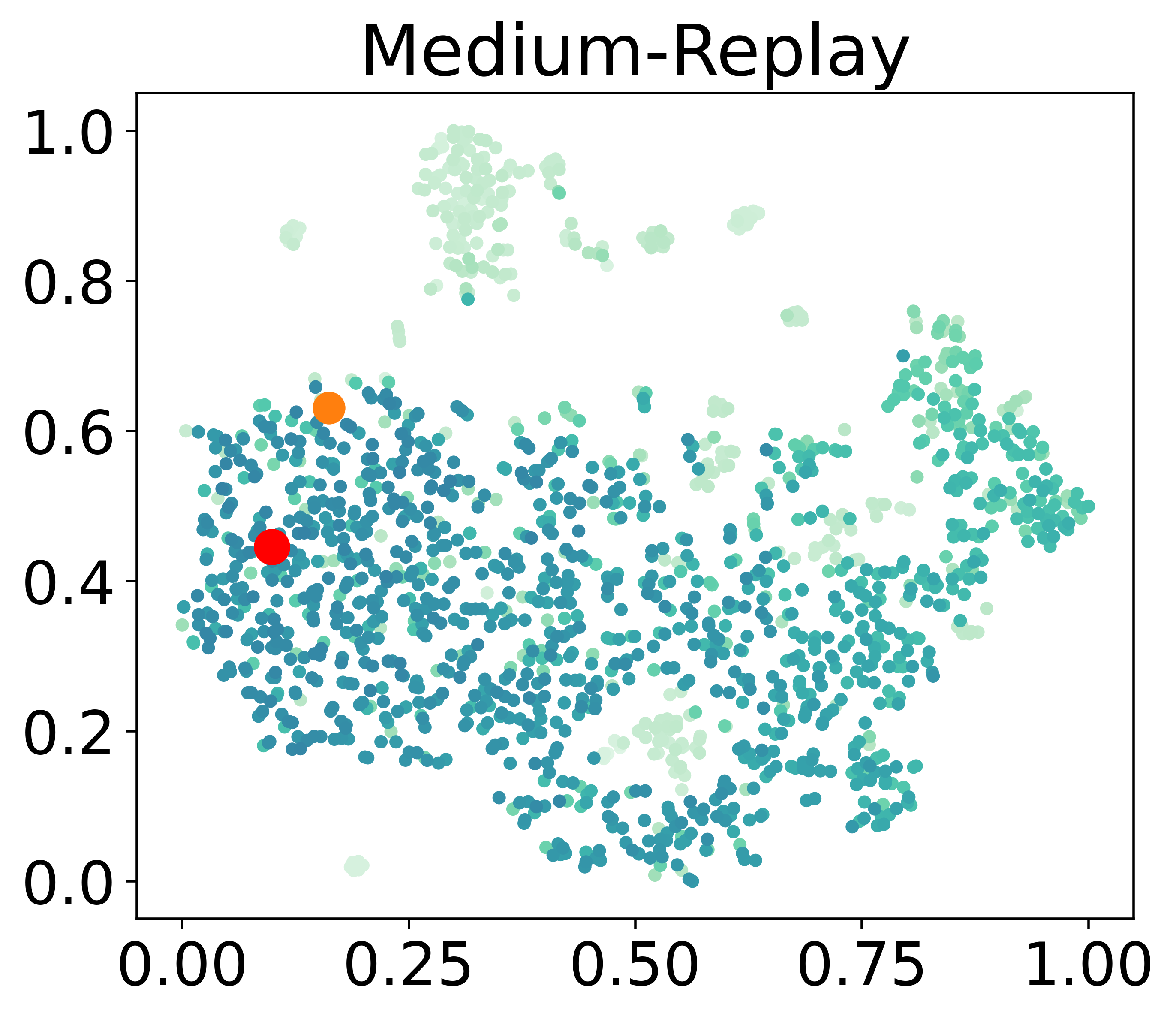}
                                \includegraphics[width=0.29\textwidth]{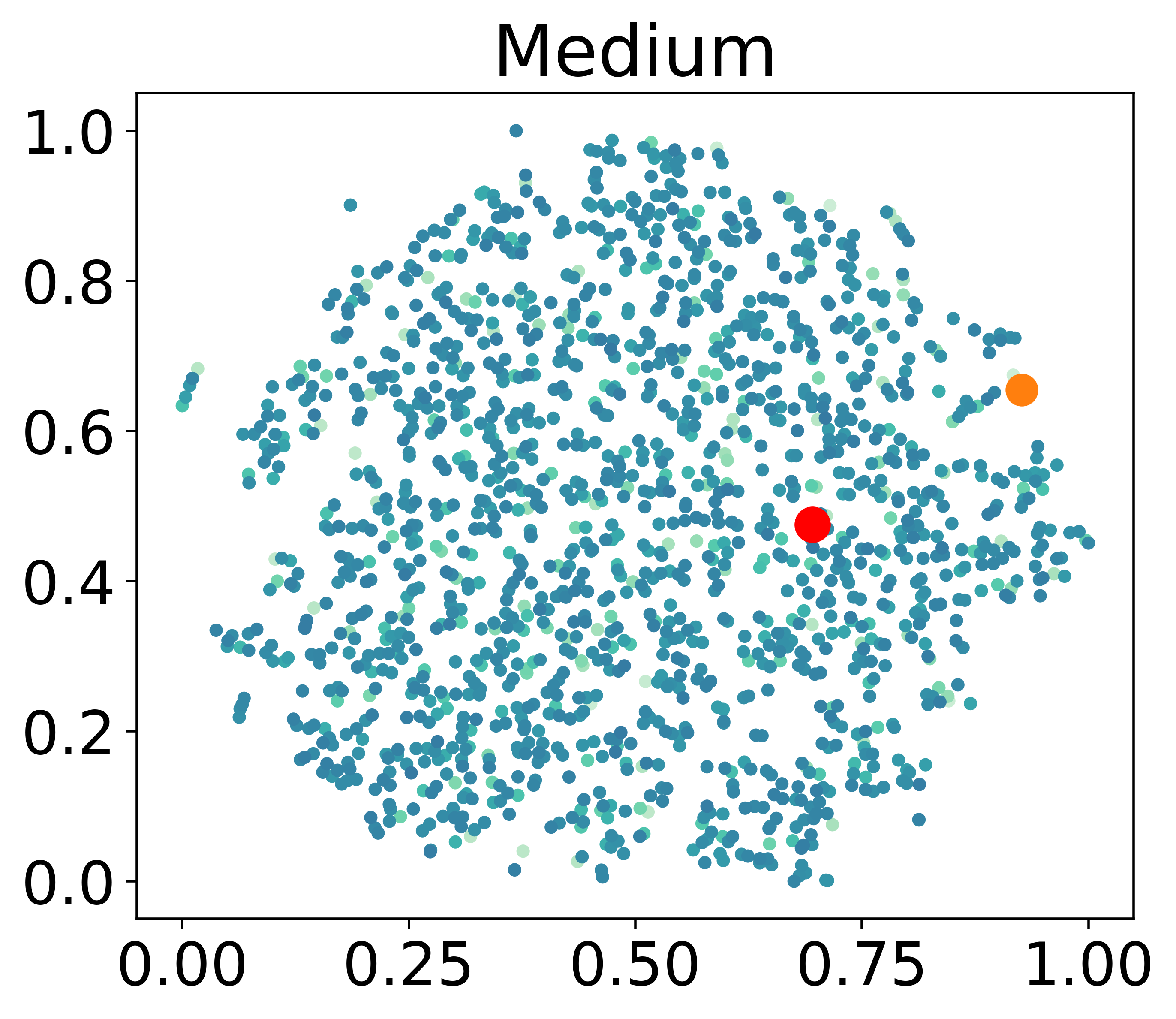}
                                \includegraphics[width=0.29\textwidth]{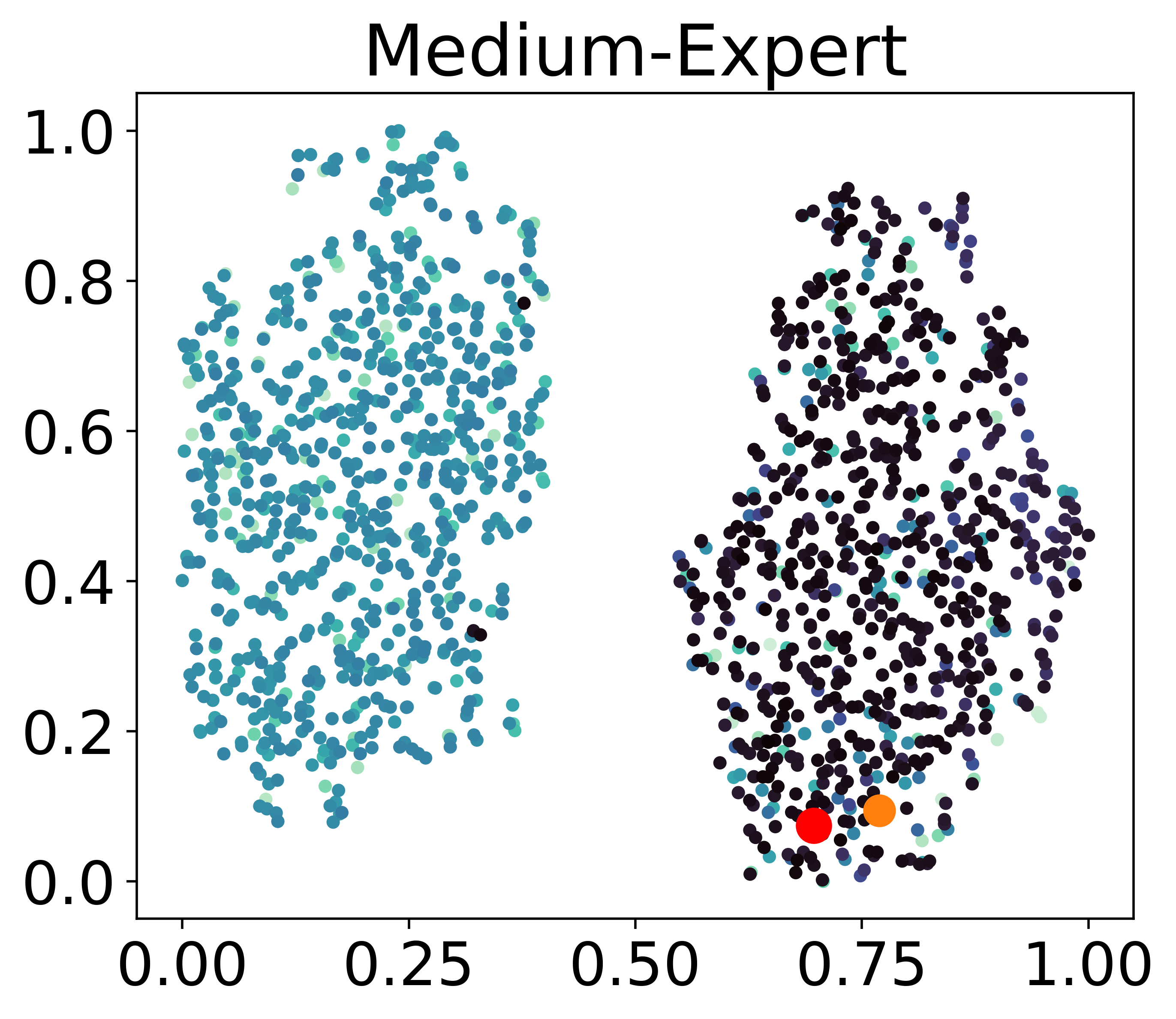}
                                \includegraphics[width=0.07\textwidth]{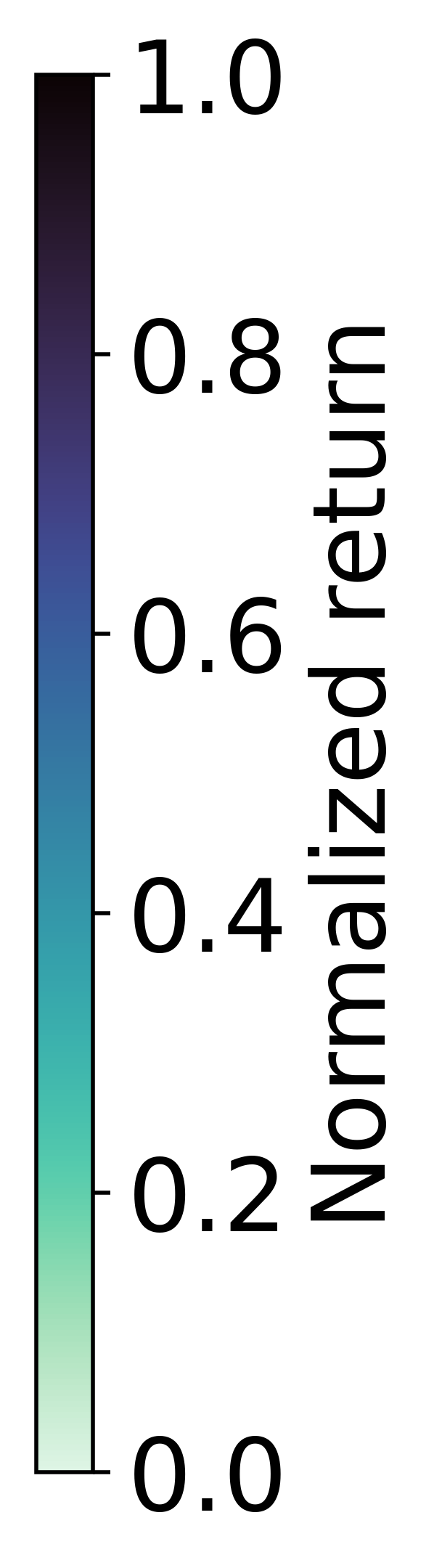}   }
        % \vspace{-5pt}
	\caption{
	    t-SNE visualization of the embedding space learned with OPPO-a in walker and halfcheetah environments.
	}
	\label{fig:tsne2-appendix}
% 	% \vspace{-10pt}
\end{figure}

We also show the t-sne results of the corresponding ablation study in \ref{fig:tsne2-appendix} with the setting described in Section \ref{sec:one-step} in Walker and HalfCheetah environments.

By comparing Fig.\ref{fig:tsne2-appendix} to Fig.\ref{fig:tsne1-appendix}, we discover that the structure of $\bz$-space significantly collapses in eight out of nine environments (except for halfcheetah medium-replay).
More specifically, we can no longer recognize the distribution pattern and clusters that emerged in Fig.\ref{fig:tsne1-appendix}, while such an observation is in line with our conclusion in the main text.

% Fundamentally, $\mathcal{L}_{\text{PM}}$ helps the encoder to learn a more informative and visually interpretable $\bz$-space.
% Removing the preference loss leads to a $\bz$-space in which the embeddings are evenly distributed in a mixing style.

\begin{table}[ht]
    \centering
    \caption{Ablation study of one-step paradigm}
    \label{tab:ablation-appendix}
    \begin{tabular}{llrrrr}
    \toprule
    \multicolumn{1}{c}{\textbf{Environment}} & \multicolumn{1}{c}{\textbf{Dataset}} & \multicolumn{1}{c}{\textbf{OPPO}}           & \multicolumn{1}{c}{\textbf{OPPO-a}}     \\ \midrule
    \multirow{3}[0]{*}{Hopper}               & Medium-Expert                        & \textbf{108.0 $\pm$ 5.1}                    & 103.5 $\pm$ 4.4                         \\
                                             & Medium                               & \textbf{86.3  $\pm$ 3.2}                    & 69.2  $\pm$ 7.4                         \\
                                             & Medium-Replay                        & \textbf{88.9  $\pm$ 2.3}                    & 78.3  $\pm$ 7.1                         \\ \midrule
    \multirow{3}[0]{*}{Walker}               & Medium-Expert                        & \textbf{105.0 $\pm$ 2.4}                    & \textbf{108.8 $\pm$ 1.0}                \\
                                             & Medium                               & \textbf{85.0  $\pm$ 2.9}                    & 80.7  $\pm$ 1.5                         \\
                                             & Medium-Replay                        & \textbf{71.7  $\pm$ 4.4}                    & 66.3  $\pm$ 1.6                         \\ \midrule
    \multirow{3}[0]{*}{HalfCheetah}          & Medium-Expert                        & \textbf{89.6  $\pm$ 0.8}                    & \textbf{90.1  $\pm$ 1.4}                \\
                                             & Medium                               & \textbf{43.4  $\pm$ 0.2}                    & \textbf{43.4  $\pm$ 0.2}                \\
                                             & Medium-Replay                        & \textbf{39.8  $\pm$ 0.2}                    & \textbf{39.6  $\pm$ 0.1}                \\ \midrule
    \multicolumn{2}{l}{\textbf{Sum}}                                                & \textbf{717.7}                              & 679.8                                   \\ \bottomrule
    \end{tabular}
\end{table}

However, it is also worth noting that the performance of OPPO-a in the D4RL benchmark is not hindered much by this uninformative $\bz$-space, as shown in Table \ref{tab:ablation-appendix}.
%We find they possess no significant performance degradation with respect to their preference-loss-enabled counterpart.
We attribute this to the effectiveness of the preference modeling phase, where our method is still able to find a meaningful $\bz^*$ in a less expressive $\bz$-space.

This is also justified from t-SNE(Fig.\ref{fig:tsne2-appendix}) as there our learned $\bz^*$ (orange dot) locates just in the point of deep color.

\end{document}